\documentclass[lettersize,journal]{IEEEtran}
\usepackage{amsmath,amsfonts}
\usepackage{algorithmic}
\usepackage{algorithm}
\usepackage{array}
\usepackage{subcaption}
\usepackage{textcomp}
\usepackage{stfloats}
\usepackage{url}
\usepackage{verbatim}
\usepackage{graphicx}
\usepackage{cite}
\usepackage{hyperref}
\usepackage{cleveref}
\usepackage{mathtools}
\usepackage{multirow, multicol, makecell}  
\usepackage{soul, color, xcolor}
\usepackage{utfsym}
\usepackage{booktabs}

\begin{document}

\title{ChatStitch: Visualizing Through Structures via Surround-View Unsupervised Deep Image Stitching with Collaborative LLM-Agents}

\author{Hao Liang, Zhipeng Dong, Kaixin Chen, Jiyuan Guo, Yufeng Yue, Mengyin Fu, Yi Yang$^{*}$
\thanks{*This work was supported by the National Natural Science Foundation of China under Grant 62233002, Grant 92370203 and Development Program of China under Grant 2022YFC2603600.(Corresponding author: Yi Yang)}
\thanks{The authors are with School of Automation, Beijing Institute of Technology, Beijing, 100081, China.}
}

\markboth{Journal of \LaTeX\ Class Files,~Vol.~14, No.~8, August~2021}%
{Shell \MakeLowercase{\textit{et al.}}: A Sample Article Using IEEEtran.cls for IEEE Journals}


\maketitle

\begin{abstract}

Surround-view perception has garnered significant attention for its ability to enhance the perception capabilities of autonomous driving vehicles through the exchange of information with surrounding cameras. However, existing surround-view perception systems are limited by inefficiencies in unidirectional interaction pattern with human and distortions in overlapping regions exponentially propagating into non-overlapping areas. To address these challenges, this paper introduces ChatStitch, a surround-view human-machine co-perception system capable of unveiling obscured blind spot information through natural language commands integrated with external digital assets. To dismantle the unidirectional interaction bottleneck, ChatStitch implements a cognitively grounded closed-loop interaction multi-agent framework based on Large Language Models. To suppress distortion propagation across overlapping boundaries, ChatStitch proposes SV-UDIS, a surround-view unsupervised deep image stitching method under the non-global-overlapping condition. We conducted extensive experiments on the UDIS-D, MCOV-SLAM open datasets, and our real-world dataset. Specifically, our SV-UDIS method achieves state-of-the-art performance on the UDIS-D\footnote{we extract a subset of data from the original UDIS-D dataset for the multi-image stitching experiment.} dataset for 3, 4, and 5 image stitching tasks, with PSNR improvements of 9\%, 17\%, and 21\%, and SSIM improvements of 8\%, 18\%, and 26\%, respectively.The code is available at https://github.com/lhlawrence/ChatStitch.
\end{abstract}


\section{Introduction}
\IEEEPARstart{S}{urround-view} Perception plays a crucial role in autonomous driving, where its primary goal is to extract pertinent information from the surrounding environment and convey it to subsequent tasks such as mapping, planning, and control\cite{mofisslam}. With the continuous advancements in large language models(LLMs), the fusion of human-centric design and artificial intelligence capabilities has opened up new possibilities for next-generation autonomous vehicles that go beyond traditional transportation\cite{llminautonomous}\cite{chatnav}.

\begin{figure}[t]
  \centering
   \includegraphics[width=\linewidth]{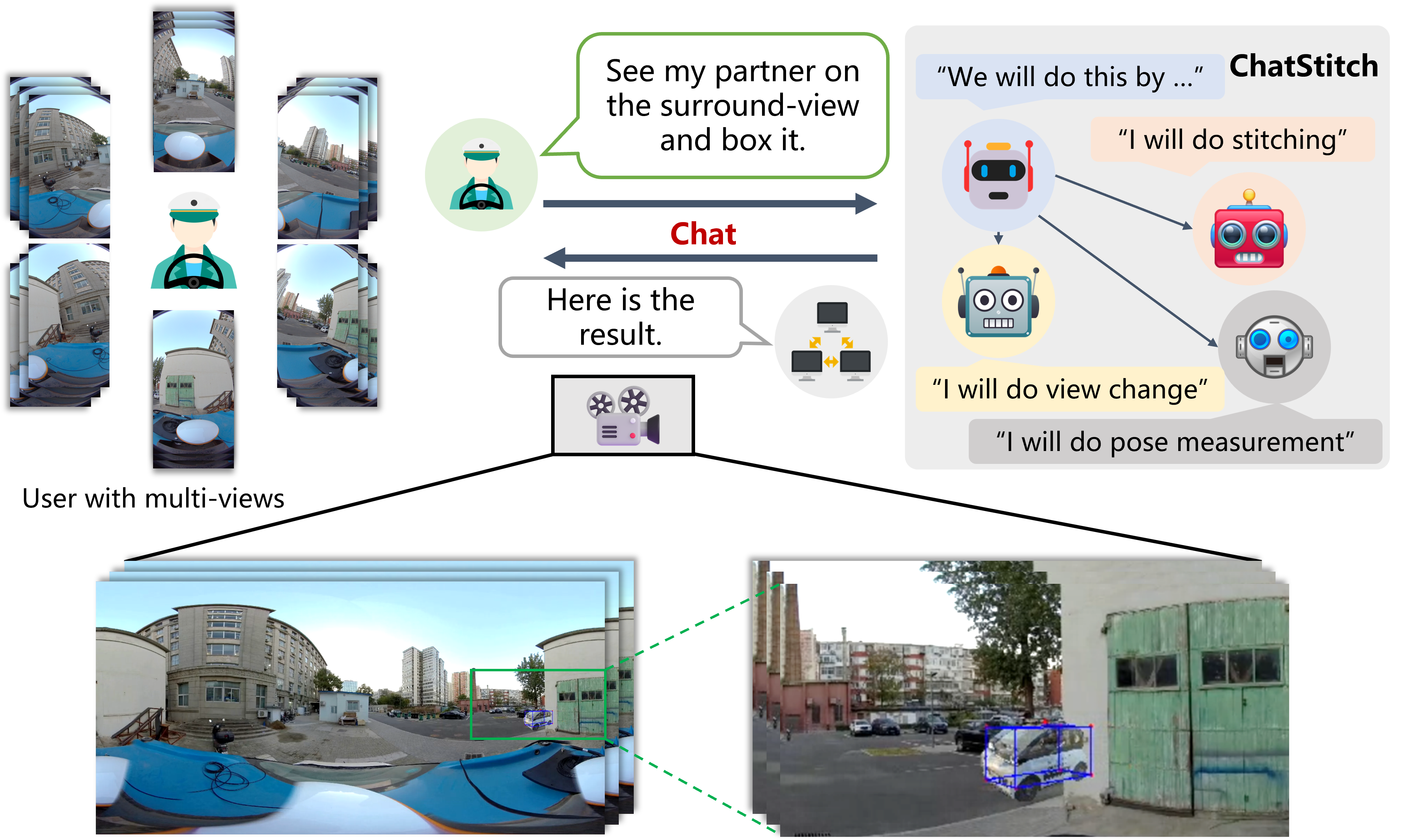}
   \caption{ChatStitch Reveals Occluded Vehicles in Stitched Surround-View Images via Language Commands.}
   \label{fig:chatstitch-abstract}
\end{figure}

\begin{figure}[t]
  \centering
  \begin{subfigure}[b]{\linewidth}
    \includegraphics[width=\linewidth]{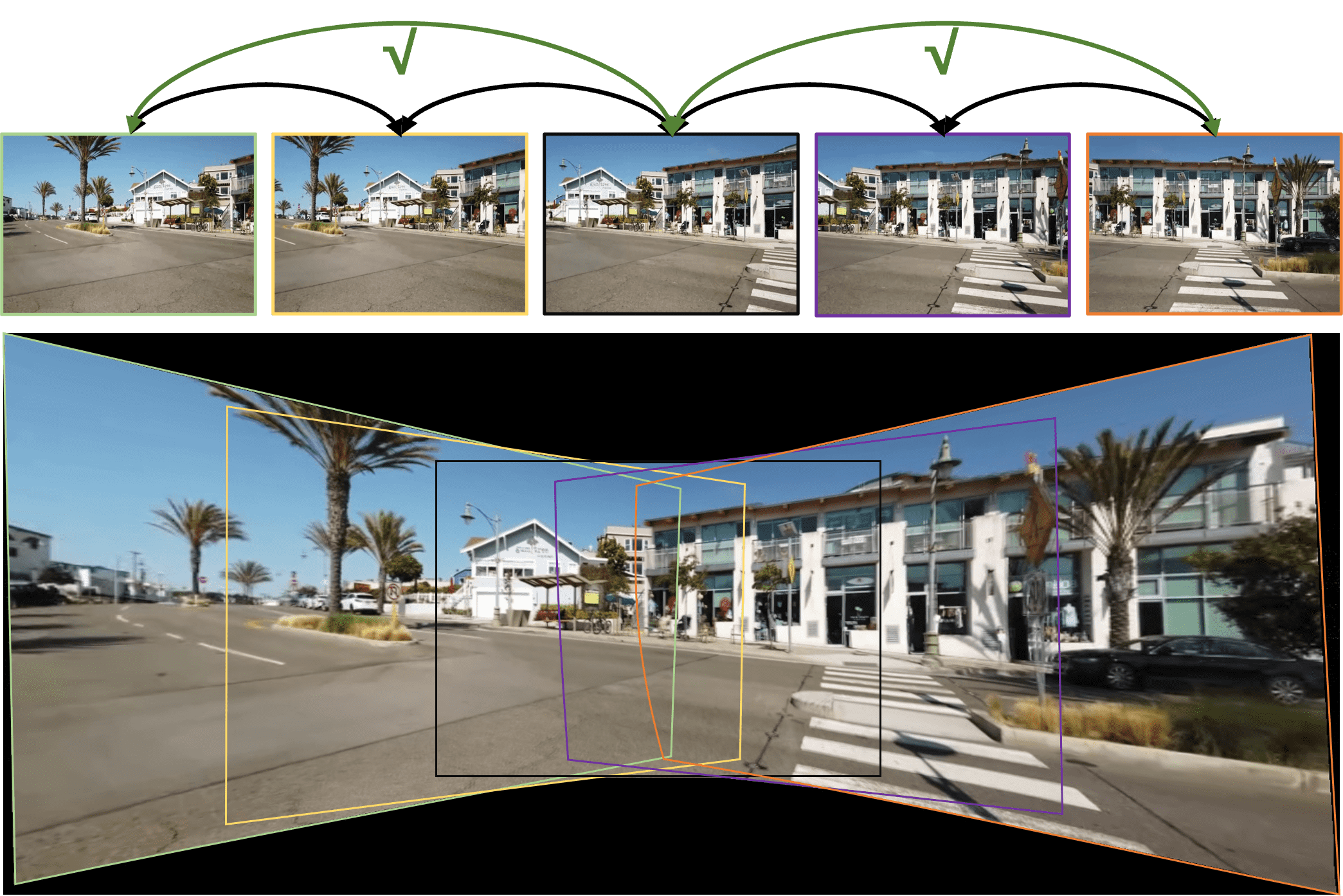}
    \caption{Global-overlapping condition and no projection}
    \label{fig:intro-stitch-a}
  \end{subfigure}
  \begin{subfigure}[b]{\linewidth}
    \includegraphics[width=\linewidth]{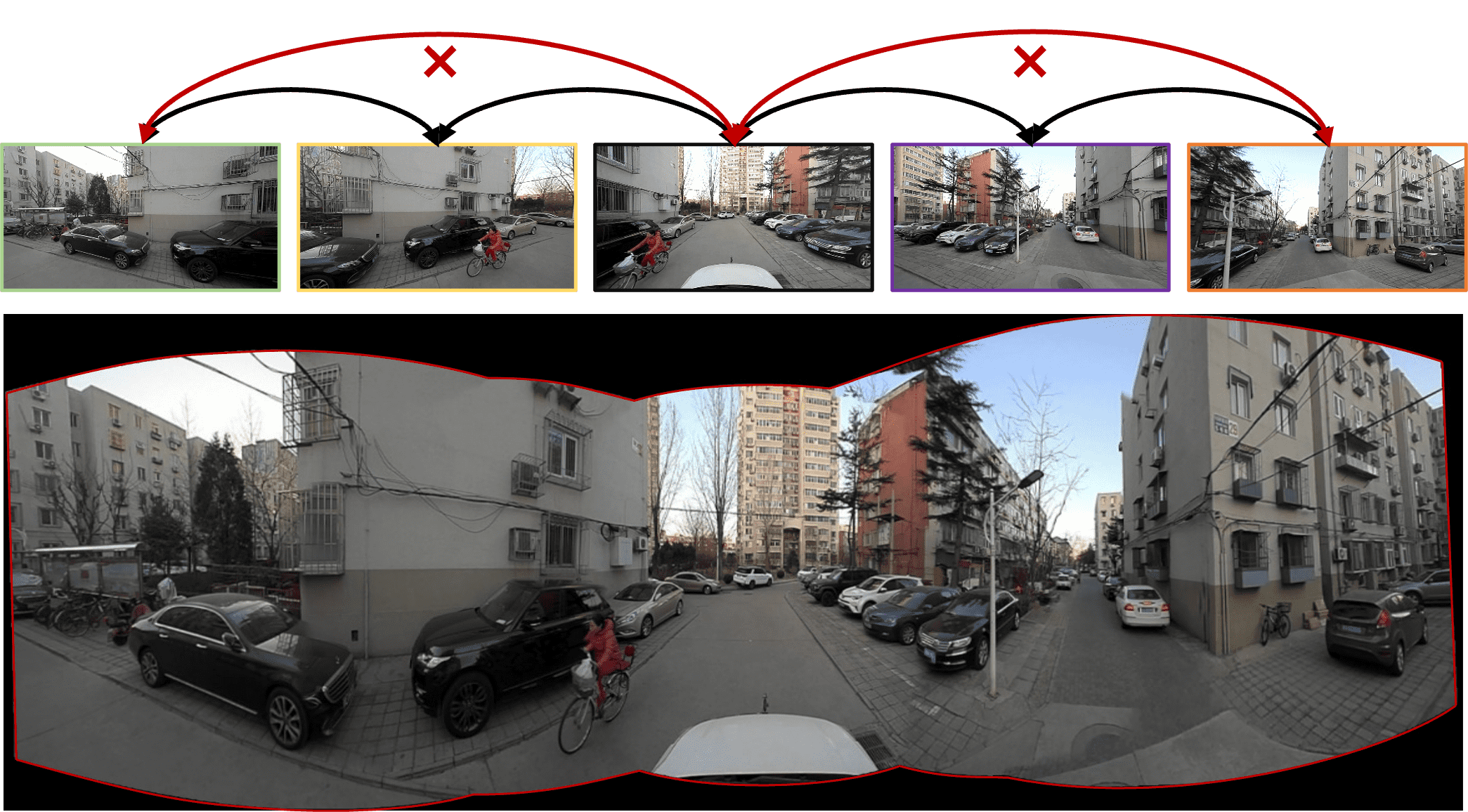}
    \caption{Non-global-overlapping condition and cylindrical projection}
    \label{fig:intro-stitch-b}
  \end{subfigure}
  \caption{Schematic illustrations of different types of multi-image stitching tasks (specific definitions in \cref{sec:sv-udis-problem-definition}) and the effects of cylindrical projection.}
  \label{fig:intro-stitch}
\end{figure}

However, existing visionLLM systems typically rely on manually preset parameters or unidirectional command inputs\cite{MCCE-REC}\cite{towards}\cite{Vision-Semantics-Label}. Interaction is limited to the initial command input, and the correction process is fully automated without iterative optimization, resulting in low interaction efficiency. To enhance system efficiency, bidirectional interaction, including real-time user feedback on correction outcomes, is necessary.
Additionally, in the context of surround-view perception, the classic task of panoramic image stitching is challenged by the constraints of vehicle layout. The overlapping regions between different images are often too small, which causes distortion propagation across these overlapping boundaries.
The current state-of-the-art (SoTA) unsupervised image stitching algorithms\cite{UDIS2, StabStitch} mainly focus on two-image stitching tasks. Although they can also perform multi-image stitching, they are not suitable for surround view images in driving scenarios.
On one hand, in these scenarios, the overlapping areas between the images to be stitched are extremely small, resulting in severe stretching at the edges of the warped images.
On the other hand, only adjacent images have overlapping areas, so there is no common reference image that has shared view areas with all other images, as shown in \cref{fig:intro-stitch-b}.
For clarity, we define the above surround-view stitching task as a multi-image stitching problem under the \textbf{non-global-overlapping} condition.
\textit{A detailed definition of this problem can be found in \cref{sec:sv-udis-problem-definition}; more discussion on  multi-image stitching please refer to our supplementary materials.}

To address these needs, we introduce ChatStitch, a surround-view human-machine co-perception system capable of unveiling obscured blind spot information through natural language commands integrated with external digital assets. To utilize ChatStitch, users simply engage in dialogue with the system, issuing commands in natural language without the need to partake in the intermediate perception steps. This process is illustrated in \cref{fig:chatstitch-abstract}.

To dismantle the unidirectional interaction bottleneck, ChatStitch implements a cognitively grounded closed-loop interaction multi-agent framework based on LLMs. The core concept revolves around utilizing multiple LLM-agents, each designated with a specific role, to decompose the overall surround-view human-machine co-perception perception requirements into distinct tasks. This reflects the functional segmentation typically established within technological frameworks. This workflow offers two primary advantages. Firstly, the Large Language Models are capable of processing human natural language commands, allowing for intuitive and dynamic adjustments to the perception displays, thereby achieving precise tuning and feedback. Secondly, the closed-loop framework enhances the effectiveness and accuracy of perception by allocating specific tasks among specialized agents, thus expanding the boundaries of surround-view perception capabilities.

To suppress distortion propagation across overlapping boundaries, We propose SV-UDIS within ChatStitch, a technical module corresponding to an agent in the ChatStitch framework, designed to stitch multiple surround-view images captured in driving scenarios into a seamless panoramic image. Our proposed  SV-UDIS method, through masked cylindrical projection, rectangular warping constraints, and motion propagation, achieves multi-image stitching under the non-global overlapping condition, providing humans with intuitive visual effects.

To demonstrate the visual advantages of surround-view perception in intuitive scenarios such as occluded buildings, we conducted experiments on UDIS-D\cite{UDIS} dataset; MCOV-SLAM\cite{PanMiaoxin} dataset and our real-world dataset. The results indicate that ChatStitch can generate photorealistic perception outcomes through occluded buildings responding to a variety of human language commands.Our proposed SV-UDIS achieves SoTA performance on the UDIS-D dataset for 3, 4, and 5 image stitching tasks, with PSNR improvements of 9$\%$, 17$\%$, and 21$\%$, and SSIM improvements of 8$\%$, 18$\%$, and 26$\%$, respectively.

In brief, we have made the following contributions:
\begin{itemize}
    \item We develop a cognitively grounded closed-loop interaction multi-agent framework, named ChatStitch, which is a surround-view human-machine co-perception system capable of unveiling obscured blind spot information through natural language commands integrated with external digital assets.
    \item We propose SV-UDIS within ChatStitch, a surround-view unsupervised deep image stitching method under the non-global-overlapping condition, which stitches common surround images in driving scenarios into a panorama, providing intuitive visual effects.
    \item We achieved photorealistic perspective effects in challenging scenarios, like occluded buildings, on both public and real-world datasets. Our method achieves SoTA performance on UDIS-D and MOCV-SLAM open datasets, resulting in a significant improvement in performance.
\end{itemize}

\section{Related work}
\label{sec:related}

\textbf{Unsupervised image stitching.} 
Image stitching technology\cite{Stitch-Survey} combines multiple images with overlapping areas to generate an image with a larger field of view.
In recent years, learning-based image stitching methods\cite{Deep-Stitch-Survey} have emerged.
Compared to traditional methods\cite{AutoStitch, APAP-TPAMI, LPC, GES-GSP} based on sparse geometric features\cite{ORB, SuperGlue, LSD}, learning-based stitching methods have shown better robustness and faster stitching speed, as they eliminate the endless design of geometric features.
Among them, due to the difficulty in obtaining real stitched labels, unsupervised stitching methods\cite{UDIS, UDIS2, SuperUDIS, FastUDIS} represented by UDIS\cite{UDIS} and UDIS2\cite{UDIS2} are more popular than supervised methods\cite{Superviesed-VFISNet, Supervised-Weekly, Supervised-SPL, Supervised-Edge}.
However, few have considered multi-image stitching under the non-global-overlapping condition, which is common in driving scenarios, such as the camera configurations used in the NuScenes\cite{NuScenes}, Waymo\cite{Waymo} and PandaSet\cite{PandaSet} datasets, where all images are horizontally arranged with extremely low overlapping areas.
In this paper, we propose SV-UDIS, a multi-image stitching algorithm for non-global-overlapping situations.

\textbf{Large Language Model and multi-agent framework.} 
Large Language Models are designed to integrate vast amounts of data for understanding, generating, and responding to human language. Since Google introduced the Transformer model \cite{Attention_is_all_you_need}, a plethora of models and methods based on it have emerged, particularly exemplified by the Large Language Models such as GPT\cite{GPT4o} and BERT\cite{bert}. Building on LLMs, many efforts have begun utilizing rich multi-agent collaborative pattern components\cite{wei2024editable}, enabling agents to function optimally in their respective roles across various domains. In our work, we explore the use of a multi-agent framework to accomplish surround-view perception tasks.

\textbf{Collaborative perception.} 
Existing collaborative perception methods are primarily categorized into three main types\cite{han2023collaborative}: Early Collaboration\cite{chen2019cooper, arnold2020cooperative, li2021learning}, Intermediate Collaboration\cite{liu2020when2com, wang2020v2vnet, xu2022v2x, xu2022cobevt, hu2023collaboration}, and Late Collaboration\cite{yuan2022keypoints, yu2022dair}. 
A comparison of different methods is shown in \cref{tab:comparison_collaborative_perception}.
However, regardless of the type of collaborative perception method employed, most focus primarily on the interactions between agents and the final task outcomes, lacking human participation and intuitive perception in collaborative perception methods. In our work, Chatstitch, adhering to the philosophy of human-in-the-loop, enables the adjustment and intuitive display of perception tasks through the processing of human natural language.

\begin{figure*}[t]
  \centering
   \includegraphics[width=\linewidth]{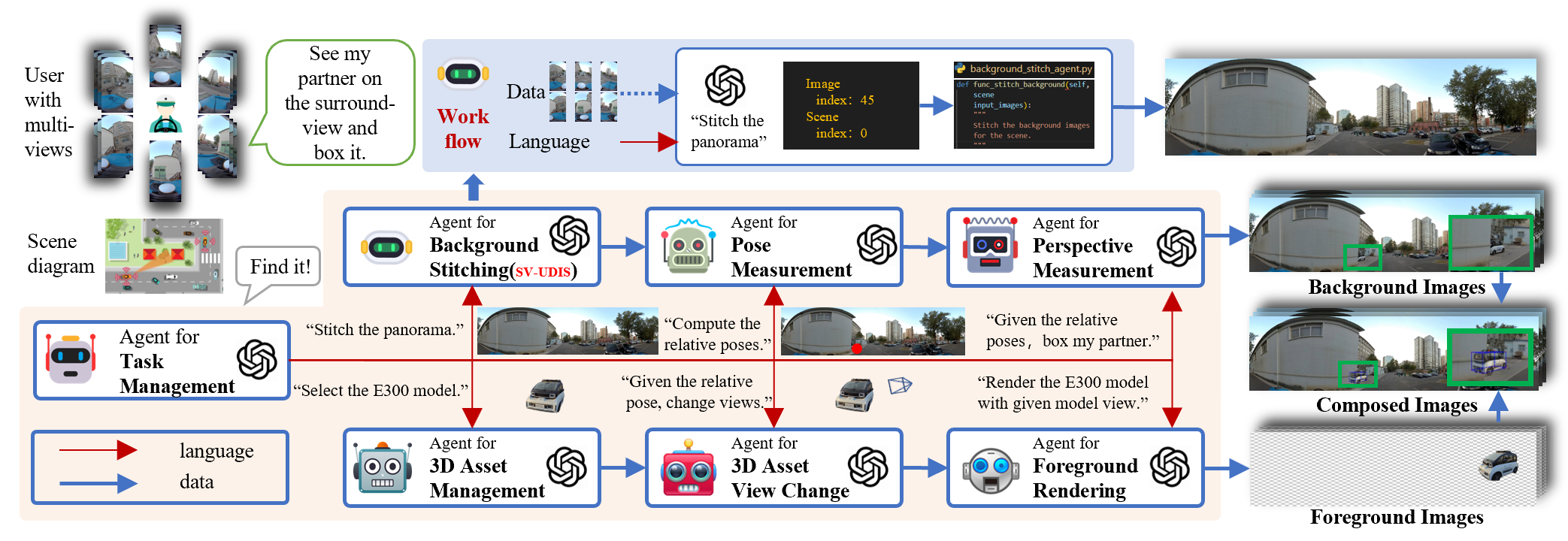}
   \caption{\textbf{Overview of the ChatStitch System.} ChatStitch breaks down complex human language into segments for different agents, each equipped with its own language processing model and executable functions.}
   \label{fig:overview-chatstitch}
\end{figure*}

\begin{table}[t]
    \centering
    \caption{Comparison of existing and proposed methods for collaborative perception.}
    \resizebox{\columnwidth}{!}{
    \setlength{\tabcolsep}{1.0mm}{
    \begin{tabular}{c|cccccccc}
        \toprule
        Method  & \makecell[c]{Photo- \\ realistic} & Dim. & \makecell[c]{Multi- \\ camera} & Editable & \makecell[c]{External \\ assets} & Language & \makecell[c]{Open- \\ source} & Stitching \\
        \midrule
        AirSim\cite{airsim2017fsr} & \usym{2613} & 3D & \usym{1F5F8} & \usym{1F5F8} & \usym{1F5F8} & \usym{2613} & \usym{1F5F8} & \usym{2613} \\
        OpenScenario\cite{OpenscanerioEditor} & \usym{2613} & 3D & \usym{1F5F8} & \usym{1F5F8} & \usym{1F5F8} & \usym{2613} & \usym{1F5F8} & \usym{2613} \\
        51Sim-One\cite{51Sim-One} & \usym{2613} & 3D & \usym{1F5F8} & \usym{1F5F8} & \usym{1F5F8} & \usym{2613} & \usym{2613} & \usym{2613} \\
        Who2com\cite{liu2020who2com} & \usym{2613} & 3D & \usym{1F5F8} & \usym{1F5F8} & \usym{1F5F8} & \usym{2613} & \usym{2613} & \usym{2613} \\
        When2com\cite{liu2020when2com} & \usym{2613} & 3D & \usym{1F5F8} & \usym{1F5F8} & \usym{1F5F8} & \usym{2613} & \usym{1F5F8} & \usym{2613} \\
        MP-Pose\cite{zhou2022multi} & \usym{2613} & 3D & \usym{1F5F8} & \usym{1F5F8} & \usym{1F5F8} & \usym{2613} & \usym{2613} & \usym{2613} \\
        CoBEVT\cite{xu2022cobevt} & \usym{2613} & 3D & \usym{1F5F8} & \usym{1F5F8} & \usym{1F5F8} & \usym{2613} & \usym{1F5F8} & \usym{2613} \\
        CoCa3D\cite{hu2023collaboration} & \usym{2613} & 3D & \usym{1F5F8} & \usym{1F5F8} & \usym{1F5F8} & \usym{2613} & \usym{1F5F8} & \usym{2613} \\
        VIMI\cite{wang2023vimi} & \usym{2613} & 3D & \usym{1F5F8} & \usym{1F5F8} & \usym{1F5F8} & \usym{2613} & \usym{2613} & \usym{2613} \\
        HM-ViT\cite{xiang2023hm} & \usym{2613} & 3D & \usym{1F5F8} & \usym{1F5F8} & \usym{1F5F8} & \usym{2613} & \usym{2613} & \usym{2613} \\
        P-CNN\cite{xiong2021toward} & \usym{1F5F8} & 2D & \usym{1F5F8} & \usym{2613} & \usym{2613} & \usym{2613} & \usym{2613} & \usym{2613} \\
        BEVGen\cite{swerdlow2024street} & \usym{1F5F8} & 2D & \usym{1F5F8} & \usym{1F5F8} & \usym{2613} & \usym{2613} & \usym{1F5F8} & \usym{2613} \\
        BEVControl\cite{yang2023bevcontrol} & \usym{1F5F8} & 2D & \usym{1F5F8} & \usym{1F5F8} & \usym{2613} & \usym{2613} & \usym{2613} & \usym{2613} \\
        DriveDreamer\cite{wang2023drivedreamer} & \usym{1F5F8} & 2D & \usym{1F5F8} & \usym{1F5F8} & \usym{2613} & \usym{1F5F8} & \usym{2613} & \usym{2613} \\
        DrivingDiffusion\cite{li2023drivingdiffusion} & \usym{1F5F8} & 2D & \usym{1F5F8} & \usym{1F5F8} & \usym{2613} & \usym{1F5F8} & \usym{2613} & \usym{2613} \\
        P2OD\cite{bi2022achieving} & \usym{1F5F8} & 2D & \usym{1F5F8} & \usym{2613} & \usym{2613} & \usym{2613} & \usym{2613} & \usym{2613} \\
        MagicDrive\cite{gao2023magicdrive} & \usym{1F5F8} & 2D & \usym{1F5F8} & \usym{1F5F8} & \usym{2613} & \usym{2613} & \usym{2613} & \usym{2613} \\
        UniSim\cite{yang2023unisim} & \usym{1F5F8} & 3D & \usym{2613} & \usym{1F5F8} & \usym{2613} & \usym{2613} & \usym{2613} & \usym{2613} \\
        MARS\cite{wu2023mars} & \usym{1F5F8} & 3D & \usym{2613} & \usym{1F5F8} & \usym{2613} & \usym{2613} & \usym{1F5F8} & \usym{2613} \\
        ChatSim\cite{wei2024editable} & \usym{1F5F8} & 3D & \usym{1F5F8} & \usym{1F5F8} & \usym{1F5F8} & \usym{1F5F8} & \usym{1F5F8} & \usym{2613} \\
        \textbf{ChatStitch(Ours)} & \usym{1F5F8} & 3D & \usym{1F5F8} & \usym{1F5F8} & \usym{1F5F8} & \usym{1F5F8} & \usym{1F5F8} & \usym{1F5F8} \\
        \bottomrule
    \end{tabular}
        }
    }
    \label{tab:comparison_collaborative_perception}
\end{table}

\section{Multi-Agent Framework}
\label{sec:framework}

ChatStitch analyzes complex and abstract commands in human-machine co-perception system, providing users with surround-view perception visualization videos. Please refer to \cref{fig:overview-chatstitch}. 
Given the complexity of human language and the abstract nature of human-machine co-perception tasks, our system is designed to flexibly handle specific aspects or multiple components of these tasks. 
The challenge of directly applying a single LLM-agent lies in its struggle with multi-step reasoning and cross-referencing \cite{wei2024editable}. Drawing from this insight, we employ a multi-agent framework with LLM agents, where each agent is responsible for managing a specific module of the perception task.

\subsection{Agents' Functionality}
\label{sec:agents}

In our framework, agents have two primary functions: On one hand, they receive concise, individual instructions from the Task Management Agent, which have been decomposed from larger tasks, and convert these instructions into corresponding parameters using predefined LLM settings. On the other hand, they input these transformed parameters into specific functions to execute the related operations. In other words, each agent is a composite of a specific LLM and corresponding functional modules. This structure not only allows the agents to process human language instructions but also to perform precise operations accordingly. \textit{For an example of how an agent converts parameters, please refer to the supplementary materials.}

\textbf{Task Management Agent.} The Task Management Agent converts complex and abstract human commands, expressed in sophisticated language, into multiple concise task statements. This agent's specialized LLM module is designed to decompose these intricate commands into a combination of specific tasks. The corresponding specific function of this agent then channels these output sentences into relevant agents according to their interrelationships.

\textbf{Background Stitching Agent.} The Background Stitching Agent seamlessly stitches vehicle images into a panoramic view, providing users with a 360° surround visual field. This agent's dedicated LLM module receives stitching instructions and executes the corresponding specific function. Notably, within this agent, we introduce a novel stitching method SV-UDIS, which addresses the issue of multi-image stitching under the non-global-overlapping condition. \textit{For detailed information, please refer to \cref{sec:sv-udis}. }

\textbf{Relative Pose Measurement.} The Relative Pose Measurement Agent transforms the positions of partners into its own coordinate system and marks them on the panoramic view. This agent's specialized LLM module receives the IDs and corresponding GPS positions of all vehicles, converting them into the appropriate parameters. Its corresponding specific function then translates the partner positions into its own image coordinate system and marks the respective locations on the panoramic view.

\textbf{Perspective Measurement.} The Perspective Measurement Agent translates 3D partner dimensions into panoramic view representations, marking bounding boxes accordingly. Its specialized LLM module processes vehicle IDs and dimensions, converting them into parameters for perspective transformations. This function maps dimensions onto the image coordinate system, delineating bounding boxes on the panoramic view.

\textbf{3D Asset Management.} The 3D Asset Management Agent selects and manages partner 3D models based on specific instructions. Its LLM module identifies relevant models from the 3D Asset Bank and outputs their IDs, enabling modification and replacement as needed. \textit{For a more comprehensive introduction to the 3D Asset Bank, please refer to the supplementary materials.}

\textbf{3D Asset View Change.} The 3D Asset View Change Agent determines external view parameters based on partners' relative positions. Its LLM module processes vehicle IDs and orientations, converting them into parameters. The function then applies coordinate transformations to align with the model's system and stores the respective viewpoints.

\textbf{Foreground Rendering.} The Foreground Rendering Agent combines viewpoint data and 3D asset information to render partners in a surround-view perception scenario. Its LLM module receives vehicle IDs and activates a specific function, which uses these parameters to call the instant-ngp interface\cite{instant} for rendering partners' images.

\subsection{Workflow}

\begin{figure*}[ht]
  \centering
   \includegraphics[width=\linewidth]{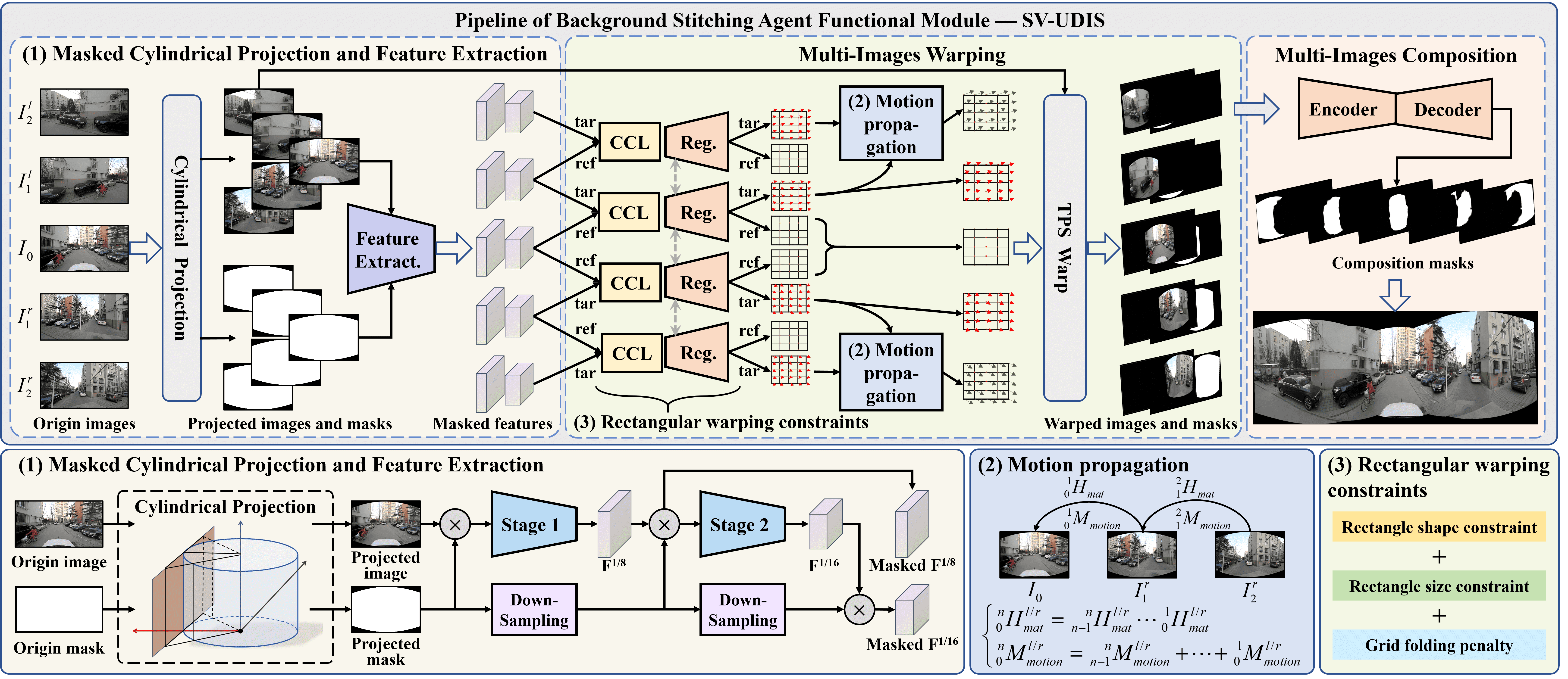}
   \caption{\textbf{Overview of our proposed SV-UDIS.} The pipeline mainly includes three stages: Masked cylindrical projection and feature extraction, multi-image warping, and multi-image composition. Our main contributions are shown in detail at the bottom of the figure.}
   \label{fig:sv-udis-framework}
\end{figure*}

ChatStich is composed of multiple agents that collaboratively produce surround-view images of a scene based on human instructions. The Task Management Agent breaks down complex human instructions into simpler task directives and delivers them to subsequent other agents. These agents work in two separate processes: Background Stitching and Foreground Rendering.

In the Background Stitching process, the initial step involves stitching images from cameras surrounding the vehicle to create a panoramic view. Subsequently, relative positions and perspective scaling ratios are calculated based on the coordinates of collaborative units, and these are marked on the panoramic image.

In the Foreground Rendering process, the preferred models from the 3D Asset Bank are selected based on prior knowledge. The rendering angles are then calculated based on the poses, and interfaces are called to visually represent the collaborators effectively.

The final images from Foreground Rendering and Background Stitching are integrated to achieve a photo-realistic result for surround-view perception. Thanks to the Task Management Agent's ability to store instructions and parameters for other agents, it enables scene modifications through multiple rounds of instructions, enhancing dynamic interaction and responsiveness in the scene. The closed-loop framework enhances the effectiveness and accuracy of perception by allocating specific tasks among specialized agents, thus expanding the boundaries of surround-view perception capabilities.

\section{SV-UDIS within Background Stitching Agent}
\label{sec:sv-udis}

In this section, We propose SV-UDIS within ChatStitch, a technical module corresponding to an agent in the ChatStitch framework, designed to stitch multiple images captured in driving scenarios into a seamless panoramic image under the non-global-overlapping condition. we first describe the definition of the non-global-overlapping condition, and then introduce the overall framework of SV-UDIS, as shown in \cref{fig:sv-udis-framework}. 

\subsection{Problem Definition}
\label{sec:sv-udis-problem-definition}

In this paper, we aim to stitch $N$ adjacent images captured by the surround-view cameras into a panoramic image, with the camera configuration identical to that of MCOV-SLAM\cite{PanMiaoxin}.
For a set of surround view image data $\{I_i\}_{i=1,...,N}, N>3$ with the resolution of $W \times H$ collected at a certain moment, we define the \textbf{non-global-overlapping} condition (shown in \cref{fig:intro-stitch-b}) as: for any image $I_i$ captured by camera $i$, there is and only exists overlapping areas with images captured by adjacent cameras. Represented as:
\begin{equation}
\begin{cases}
\forall i, \text{overlap}(I_i, I_{i-1}) \wedge \text{overlap}(I_i, I_{i+1}) \\
\neg \exists j, j \neq i-1 \wedge j \neq i+1 \wedge \text{overlap}(I_i, I_j) \\
I_{-1} = I_N, I_{N+1} = I_0
\end{cases}
\label{non-global-overlapping-condition}
\end{equation}

Similarly, we define the \textbf{global-overlapping} condition (shown in \cref{fig:intro-stitch-a}) as: there exists an image ${I_k}$, with which all other images have overlapping areas. Represented as:
\begin{equation}
\exists k, \forall i \neq k, \text{overlap}(I_i, I_k)
\label{global-overlapping-condition}
\end{equation}

Our goal is to construct a self-supervised stitching method $\mathbf{S}(\cdot)$ with learnable parameters $\theta$, which takes $N$ images satisfying the non-global-overlapping condition described in \cref{non-global-overlapping-condition} as input, and stitches them to generate a panoramic image $I_{sv}$. It is defined as
\begin{equation}
\mathbf{S}(I_1, ..., I_N; \theta) \to I_{sv}
\label{total-problem}
\end{equation}

\subsection{Unsupervised Multi-Image Stitching}
\label{sec:sv-udis-details}

The proposed SV-UDIS framework, as shown in \cref{fig:sv-udis-framework}, mainly includes three stages: Masked cylindrical projection and feature extraction, multi-image warping, and multi-image composition.
In the first stage, we perform cylindrical projection on the input $N$ images, generate projected images and projected masks, and extract multi-scale features.
In the second stage, we take two adjacent images as a group, regress their features, and generate the mesh warping of the target image for each group. Then, using our proposed motion propagation strategy, we calculate the mesh motion between the edge image and the central image, thereby generating aligned images and masks through TPS transformation\cite{UDIS2}.
In the third stage, we predict the image fusion masks through a UNet-like network, thereby generating the final stitched image.

\textbf{Masked cylindrical projection and feature extraction.} 
In this paper, we aim to stitch multiple images with a large horizontal field of view. To maintain the spatial constraints and visual consistency of the images, we need to process the images with cylindrical projection, i.e., project the images from each camera onto the cylindrical surface, and then unfold the cylindrical surface to achieve the projection deformation of each image.
The impact of whether to perform cylindrical projection on the multi-image stitching task with a large horizontal field of view is shown in \cref{fig:intro-stitch}.

For any pixel point $p(x,y)$ on the input image $I_i$, it is projected into the camera coordinate system $P(X,Y,Z)$ according to the camera's internal parameters $K$, then transformed into the cylindrical latitude and longitude coordinate system $P_c(\lambda,\varphi)$, and finally unfolded to obtain the pixel coordinates $p'(u,v)$ on the projected image. The specific process is:
\begin{equation}
\begin{aligned}
\phantom{\Rightarrow} & P = K^{-1} p \\
\Rightarrow & \lambda = \arctan \frac{X}{Z}, \varphi = \arctan \frac{Y}{\sqrt{X^2+Z^2}} \\
\Rightarrow & u = \lambda = \arctan \frac{X}{Z}, v = \tan\varphi = \frac{Y}{\sqrt{X^2+Z^2}}
\end{aligned}
\label{forward-cylindrical-projection}
\end{equation}

The above is the forward projection process. However, to obtain a cylindrical projected image, it is necessary to reverse project the point $p'(u,v)$ on the cylindrical image to the corresponding pixel point $p(x,y)$ on the original image, and perform pixel interpolation to obtain a complete cylindrical image. The specific process is as follows:
\begin{equation}
\begin{aligned}
\phantom{\Rightarrow} & \lambda = u, \varphi = \arctan v \\
\Rightarrow & X = \sin\lambda, Y = v, Z = \cos\lambda \\
\Rightarrow & p = KP
\end{aligned}
\label{backword-cylindrical-projection}
\end{equation}

After performing cylindrical projection on the input image, we obtain the projected image and its mask. We use ResNet50\cite{ResNet} with pretrained parameters as the backbone network to extract high-dimensional semantic features, and successively obtain feature maps $F^{1/8}$ and $F^{1/16}$ with resolutions of 1/8 and 1/16 of the original image, respectively.
Due to the presence of some black areas in the corners of the projected image, in order to avoid learning these black areas and their contours as significant features, we apply the projected mask during the feature extraction process. As shown in \cref{fig:sv-udis-framework}, we finally obtain masked $F^{1/8}$ and $F^{1/16}$.

\textbf{Motion propagation.} 
In the second stage of the process illustrated in \cref{fig:sv-udis-framework}, we take two adjacent images as a pair, where the image closer to the center of the field of view serves as the reference image $I_r$, and the image closer to the edge of the field of view serves as the target image $I_t$.
For each pair of ref-tar images, we use the contextual correlation layer (CCL)\cite{CCL} to calculate the correlation between feature maps, and then predict the initial motion $\prescript{tar}{ref}H_{motion}$ and the residual motion $\prescript{tar}{ref}M_{motion}$ of the grid control points of the target image through regression networks\cite{UDIS2}.
For the initial motion, it represents the 4-pt parameterization of the homography warp\cite{deep_homo_estimation_arxiv}. We then convert it into a $3\times3$ matrix representation $\prescript{tar}{ref}H_{mat}$, using the normalized Direct Linear Transform (DLT) algorithm.
For the residual motion, we divide the target image evenly into a grid of $U$ rows and $V$ columns, representing the movement of $(U+1)\times(V+1)$ control points.

To stitch all images together to generate a panoramic image, it is necessary to convert the control point motion between adjacent images into global control point motion, that is, to calculate the control point motion of all images relative to the central image.
We denote the image in the center of the field of view as $I_0$, and the images on the left and right sides as $\{I_n^l\}_{n=1,...,N/2}$ and $\{I_n^r\}_{n=1,...,N/2}$, respectively.
Through our motion propagation strategy, the initial homography warp $\prescript{n}{0}H_{mat}^{l/r}$ and residual motion $\prescript{n}{0}M_{motion}^{l/r}$ of any image $I_n^{l/r}$ on the left or right side relative to the central image can be represented as:
\begin{equation}
\begin{cases}
\prescript{n}{0}H_{mat}^{l/r} = \prescript{n}{n-1}H_{mat}^{l/r} \cdots \prescript{1}{0}H_{mat}^{l/r} \\
\prescript{n}{0}M_{motion}^{l/r} = \prescript{n}{n-1}M_{motion}^{l/r} + \cdots + \prescript{1}{0}M_{motion}^{l/r}
\end{cases}
\label{motion-propagation}
\end{equation}

Through the above motion propagation strategy, we can calculate the control point motion of all images relative to the central image, thereby performing globally unified warping and alignment on all images.


\textbf{Rectangular warping constraints.}
Under the non-global-overlapping condition, the overlapping area between adjacent images is usually small. Therefore, the target image warped by the homography matrix often has severe shape stretching at the edges, as shown in \cref{fig:intro-stitch}. After motion propagation, this shape stretching at the edges is dramatically amplified, resulting in an image with severely irregular edges and scale scaling after stitching.

To mitigate the issue of shape stretching at the edges and to ensure a more natural visual effect in the stitched image, we propose the rectangular warping constraints. 
The basic idea behind these constraints is that the warped target image should be as close to a rectangle as possible, avoiding severe shape stretching and scale scaling.
We design a loss function to implement these constraints, as shown in \cref{rectangular-warping-constraints-loss} , which is composed of three parts:rectangle shape constraint $\ell_{shape}$, rectangle size constraint $\ell_{size}$, and grid folding penalty $\ell_{fold}$.
\begin{equation}
\mathcal{L}_{rectangle}^w = \gamma_1 \ell_{shape} + \gamma_2 \ell_{size} + \gamma_3 \ell_{fold} 
\label{rectangular-warping-constraints-loss}
\end{equation}


For the rectangle shape constraint, we hope that the warped grid as a whole maintains horizontal or vertical orientation. 
Denoting the sets of horizontal and vertical edges of the warped grid as $\vec{e}_{hor}$ and $\vec{e}_{ver}$, and the unit vectors in the horizontal and vertical directions as $\vec{i}$ and $\vec{j}$, respectively. 
We define this constraint as follows:
\begin{equation}
\begin{aligned}
\ell_{shape} = & \frac{1}{(U+1) \times V} \sum_{\left\{\vec{e}_{hor}\right\}} \begin{vmatrix}\langle\vec{e}, \vec{j}\rangle \cdot \frac{V}{H} \end{vmatrix} \\
+ & \frac{1}{U \times (V+1)} \sum_{\left\{\vec{e}_{ver}\right\}} \begin{vmatrix}\langle\vec{e}, \vec{i}\rangle \cdot \frac{U}{W} \end{vmatrix}
\end{aligned}
\label{rectangle-shape-constraint-loss}
\end{equation}

For the rectangle size constraint, we aim to minimize significant changes in the length of the warped grid. 
We define this constraint as follows:
\begin{equation}
\begin{aligned}
\ell_{size} = & \frac{1}{(U+1) \times V} \sum_{\left\{\vec{e}_{hor}\right\}} \Big| \begin{vmatrix}\langle\vec{e}, \vec{i}\rangle \cdot \frac{V}{W} \end{vmatrix} -1  \Big| \\
+ & \frac{1}{U \times (V+1)} \sum_{\left\{\vec{e}_{ver}\right\}} \Big| \begin{vmatrix}\langle\vec{e}, \vec{j}\rangle \cdot \frac{U}{H} \end{vmatrix} -1 \Big|
\end{aligned}
\label{rectangle-size-constraint-loss}
\end{equation}

For the grid folding penalty, the warped grid should not exhibit folding phenomena, that is, the relative positional relationship among all grid edges --- up, down, left, and right --- should remain unchanged. 
Therefore, we control the orientation of the grid edge vector $\vec{e}_{hor}$ and $\vec{e}_{ver}$ through the $RELU$ function $\sigma(\cdot)$.
It is specifically defined as follows:
\begin{equation}
\begin{aligned}
\ell_{fold} = & \frac{1}{(U+1) \times V} \sum_{\left\{\vec{e}_{hor}\right\}} \sigma ( - \langle\vec{e}, \vec{i}\rangle ) \\
+ & \frac{1}{U \times (V+1)} \sum_{\left\{\vec{e}_{ver}\right\}} \sigma ( - \langle\vec{e}, \vec{j}\rangle )
\end{aligned}
\label{grid-folding-penalty-loss}
\end{equation}

The complete image warping loss function $\mathcal{L}^w$ includes three parts: alignment constraint $\mathcal{L}_{alignment}^w$, distortion constraint $\mathcal{L}_{distortion}^w$, and the rectangular warping constraint $\mathcal{L}_{rectangle}^w$ we proposed. The alignment and distortion terms are consistent with \cite{UDIS2}.
\begin{equation}
\mathcal{L}^w = \alpha\mathcal{L}_{alignment}^w + \beta\mathcal{L}_{distortion}^w + \gamma\mathcal{L}_{rectangle}^w
\end{equation}

\textbf{Multi-Images Composition.}
We adopt the unsupervised seamless composition module proposed by UDIS2\cite{UDIS2} for image composition, and extend it to multi-image stitching. 
For the warped images and masks, we also take two adjacent images as a pair, and then predict the composition masks for each pair of images. 
At this point, except for the two images located at the edge of the field of view, each other image has two composition masks. 
Multiplying them together gives us the final composition masks shown in \cref{fig:sv-udis-framework}. 
Finally, the final stitched image can be generated using the warped images and the final composition masks.

\begin{figure}[t]
  \centering
   \includegraphics[width=\linewidth]{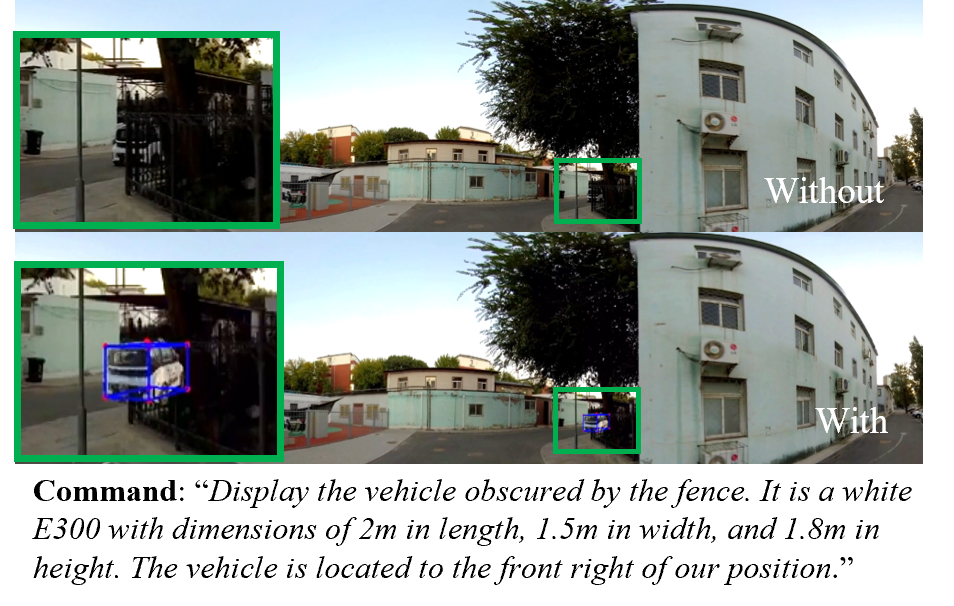}
   \caption{Result under a complex command. "With" and "without" represent the outcomes before and after our processing.}
   \label{fig:complex_command}
\end{figure}

\begin{figure}[t]
  \centering
   \includegraphics[width=\linewidth]{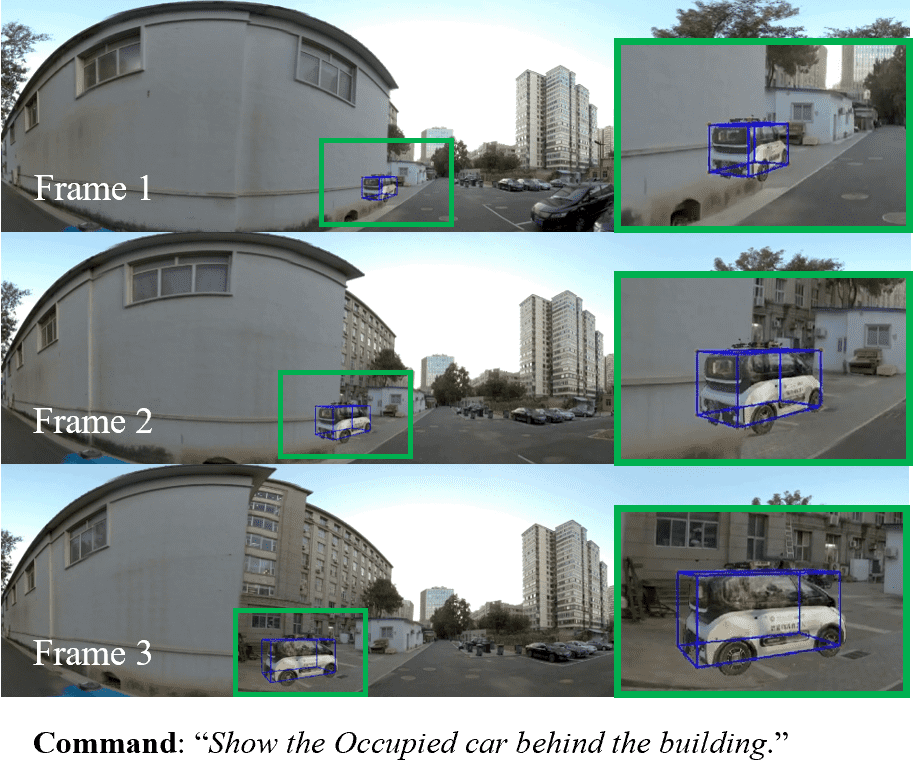}
   \caption{Result under an abstract command.}
   \label{fig:abstract_command}
\end{figure}

\section{Experiments}
\label{sec:experiments}

\subsection{Experimental settings}
\label{sec:experimental settings}
 
\textbf{ChatStitch.}
For system-level perception tasks involving Large Language Models, we conducted tests on our own real-world datasets. We aligned the positioning data from integrated navigation with panoramic image data based on GPS time, selecting a 15-second segment of typical vehicle occlusion and revelation for testing. All our Large Language Models utilized GPT-4o.

\textbf{SV-UDIS.}
For the multi-image stitching task under the non-global-overlapping condition, we train and test our model on the UDIS-D\cite{UDIS} dataset and the vehicle surround perception dataset released by MCOV-SLAM\cite{PanMiaoxin}. \textit{For more experimental settings and implementation details, please refer to our supplementary materials.}

\begin{table*}
\captionsetup{skip=3pt}
\caption{Quantitative comparison of multi-image stitching on UDIS-D$^\dag$\cite{UDIS} and MCOV-SLAM\cite{PanMiaoxin} datasets. \textcolor{red}{Red} indicates the best results. The symbol $^\dag$ denotes that we extract a subset of data from the original UDIS-D\cite{UDIS} dataset for the multi-image stitching experiment.}
\centering
\begin{subtable}{\linewidth}
\centering
\resizebox{\linewidth}{!}{
\begin{tabular}{cccccc|cccc|cccc|cccc}
\toprule
\multirow{2}*{Datasets} & \multirow{2}*{Methods} & 
    \multicolumn{4}{c}{2 images} & \multicolumn{4}{c}{3 images} & \multicolumn{4}{c}{4 images} & \multicolumn{4}{c}{5 images} \\
    \cmidrule(r){3-6} \cmidrule(r){7-10} \cmidrule(r){11-14} \cmidrule(r){15-18}
    & & Easy & Moderate & Hard & Average & Easy & Moderate & Hard & Average & Easy & Moderate & Hard & Average & Easy & Moderate & Hard & Average \\
\midrule
\multirow{3}*{\makecell[c]{UDIS\\-D$^\dag$\cite{UDIS}}} 
    & $I_{3\times3}$    & 15.87 & 12.76 & 10.68 & 12.86 & 16.87 & 12.69 & 11.35 & 13.31 & 15.08 & 12.52 & 11.36 & 12.79 & 13.66 & 12.29 & 11.40 & 12.26 \\
    & UDIS2\cite{UDIS2} & \textcolor{red}{30.19} & \textcolor{red}{25.84} & \textcolor{red}{21.57} & \textcolor{red}{25.43} & 25.18 & 22.11 & 19.12 & 21.66 & 23.21 & 20.84 & 18.20 & 20.36 & 21.53 & 20.15 & 17.57 & 19.46 \\
    & Ours              & 28.79 & 24.57 & 20.71 & 24.28 & \textcolor{red}{27.69} & \textcolor{red}{24.05} & \textcolor{red}{20.60} & \textcolor{red}{23.58} & \textcolor{red}{27.66} & \textcolor{red}{24.04} & \textcolor{red}{20.93} & \textcolor{red}{23.77} & \textcolor{red}{26.89} & \textcolor{red}{24.34} & \textcolor{red}{21.30} & \textcolor{red}{23.62} \\
\midrule
\multirow{3}*{\makecell[c]{MCOV-\\SLAM\\\cite{PanMiaoxin}}} 
    & $I_{3\times3}$    & 12.59 & 11.33 & 10.09 & 11.21 & 12.69 & 11.79 & 10.85 & 11.66 & 12.35 & 11.62 & 10.85 & 11.51 & 12.28 & 11.55 & 10.85 & 11.47 \\
    & UDIS2\cite{UDIS2} & \textcolor{red}{17.66} & \textcolor{red}{15.88} & 13.62 & 15.50 & 15.47 & 14.05 & 12.90 & 13.98 & 14.46 & 13.34 & 12.35 & 13.25 & 13.71 & 12.83 & 12.03 & 12.74 \\
    & Ours              & 17.32 & 15.79 & \textcolor{red}{14.35} & \textcolor{red}{15.67} & \textcolor{red}{16.40} & \textcolor{red}{15.33} & \textcolor{red}{14.53} & \textcolor{red}{15.31} & \textcolor{red}{16.69} & \textcolor{red}{15.85} & \textcolor{red}{14.98} & \textcolor{red}{15.73} & \textcolor{red}{16.21} & \textcolor{red}{15.48} & \textcolor{red}{14.68} & \textcolor{red}{15.36} \\
\bottomrule
\end{tabular}
}
\caption{PSNR ($\uparrow$) comparison of multi-image stitching.}  
\end{subtable}


\begin{subtable}{\linewidth}
\centering
\resizebox{\linewidth}{!}{
\begin{tabular}{cccccc|cccc|cccc|cccc}
\toprule
\multirow{2}*{Datasets} & \multirow{2}*{Methods} & 
    \multicolumn{4}{c}{2 images} & \multicolumn{4}{c}{3 images} & \multicolumn{4}{c}{4 images} & \multicolumn{4}{c}{5 images} \\
    \cmidrule(r){3-6} \cmidrule(r){7-10} \cmidrule(r){11-14} \cmidrule(r){15-18}
    & & Easy & Moderate & Hard & Average & Easy & Moderate & Hard & Average & Easy & Moderate & Hard & Average & Easy & Moderate & Hard & Average \\ 
\midrule
\multirow{3}*{\makecell[c]{UDIS\\-D$^\dag$\cite{UDIS}}} 
    & $I_{3\times3}$    & 0.530 & 0.286 & 0.146 & 0.303 & 0.572 & 0.273 & 0.157 & 0.310 & 0.459 & 0.258 & 0.157 & 0.275 & 0.355 & 0.252 & 0.164 & 0.241 \\
    & UDIS2\cite{UDIS2} & \textcolor{red}{0.933} & \textcolor{red}{0.875} & \textcolor{red}{0.739} & \textcolor{red}{0.838} & 0.848 & 0.768 & 0.624 & 0.728 & 0.779 & 0.705 & 0.573 & 0.667 & 0.719 & 0.649 & 0.550 & 0.624 \\
    & Ours              & 0.914 & 0.841 & 0.692 & 0.803 & \textcolor{red}{0.881} & \textcolor{red}{0.829} & \textcolor{red}{0.691} & \textcolor{red}{0.784} & \textcolor{red}{0.883} & \textcolor{red}{0.834} & \textcolor{red}{0.698} & \textcolor{red}{0.790} & \textcolor{red}{0.848} & \textcolor{red}{0.833} & \textcolor{red}{0.715} & \textcolor{red}{0.789} \\
\midrule
\multirow{3}*{\makecell[c]{MCOV-\\SLAM\\\cite{PanMiaoxin}}} 
    & $I_{3\times3}$    & 0.356 & 0.264 & 0.175 & 0.256 & 0.364 & 0.279 & 0.203 & 0.273 & 0.352 & 0.280 & 0.199 & 0.268 & 0.352 & 0.271 & 0.191 & 0.262 \\
    & UDIS2\cite{UDIS2} & 0.709 & 0.629 & 0.476 & 0.591 & 0.566 & 0.503 & 0.401 & 0.479 & 0.514 & 0.444 & 0.355 & 0.427 & 0.488 & 0.411 & 0.337 & 0.401 \\
    & Ours              & \textcolor{red}{0.751} & \textcolor{red}{0.694} & \textcolor{red}{0.555} & \textcolor{red}{0.655} & \textcolor{red}{0.688} & \textcolor{red}{0.634} & \textcolor{red}{0.595} & \textcolor{red}{0.634} & \textcolor{red}{0.700} & \textcolor{red}{0.656} & \textcolor{red}{0.621} & \textcolor{red}{0.655} & \textcolor{red}{0.673} & \textcolor{red}{0.624} & \textcolor{red}{0.577} & \textcolor{red}{0.619} \\
\bottomrule
\end{tabular}
}
\caption{SSIM ($\uparrow$) comparison of multi-image stitching.}  
\end{subtable}
\label{tab:comparison-multi-images}
\vspace{-15pt}
\end{table*}

\subsection{Results}
\subsubsection{System Results}

\textbf{Results with language commands.} We have selected two categories of commands, complex and abstract, to represent the most commonly used forms of human language. Complex commands consist of sentences with multiple meanings, detailed instructions, and several directives, while abstract commands are highly generalized and concise ultimate directives. To better demonstrate the perspective capabilities of ChatStitch, we have utilized images captured by an Insta360 camera to create panoramic views that serve as the background in our results.

\textit{Complex command.} We provided the system with detailed information about the type, size, and location of vehicles. \cref{fig:complex_command} displays the final results. From this, we can observe that: 1) all the information requested in the complex commands has been successfully addressed, and 2) the intuitive effects of potential collaborative targets are visible even behind obstructing buildings.

\textit{Abstract command.} We issued a highly generalized final effect command to the system. The outcome is visible in \cref{fig:abstract_command}. From this, it is evident that the highly generalized command contains minimal simple information.

\begin{figure}[t]
  \centering
   \includegraphics[width=\linewidth]{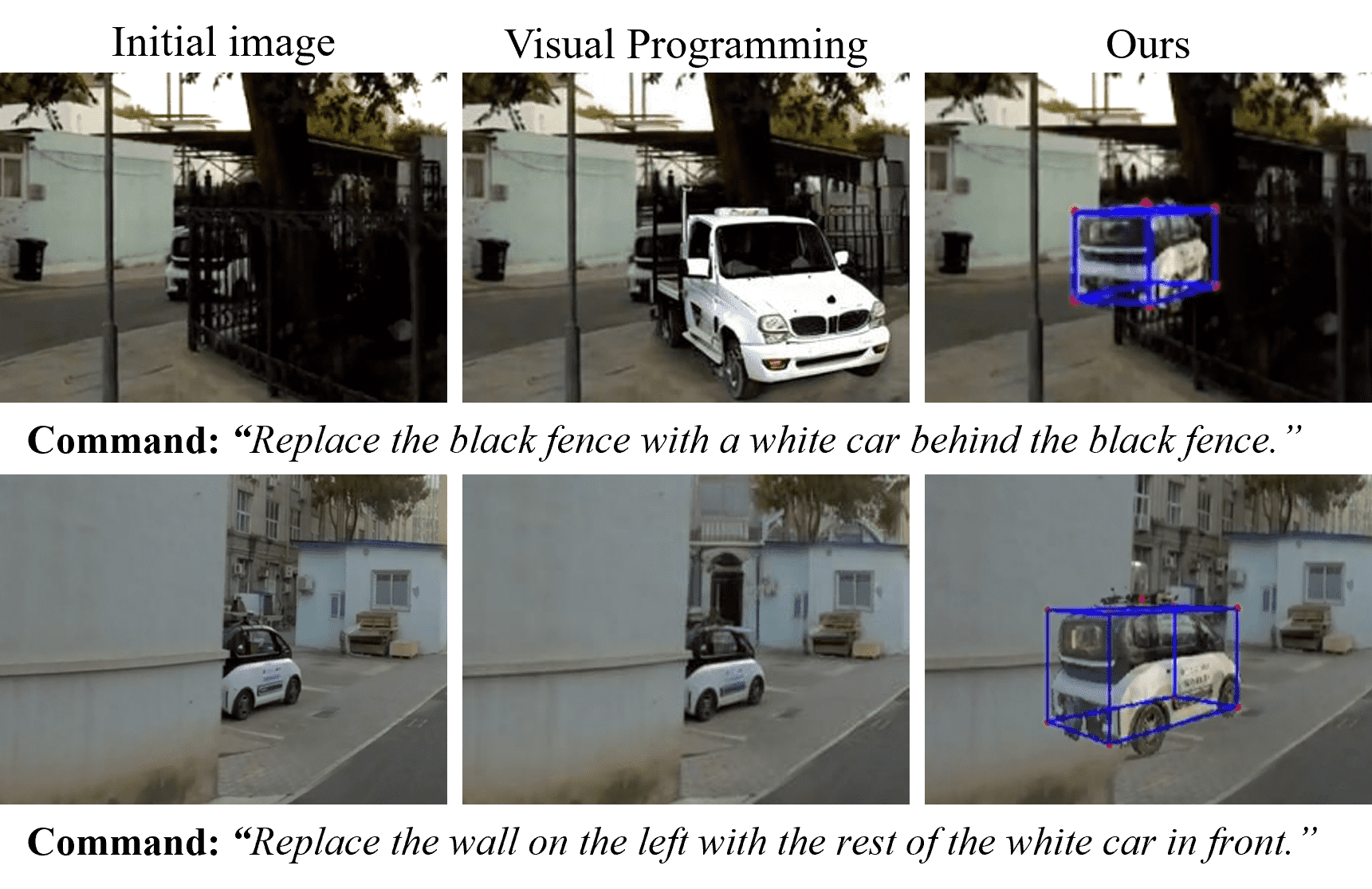}
   \caption{Qualitative comparison with Visual Programming.}
   \label{fig:comparion_visprog}
\end{figure}

\textit{Comparison with Visual Programming(VisProg)\cite{visualprograming}.} We utilized VisProg object replacement operation to remove occlusions and compared the results with our method, as shown in \cref{fig:comparion_visprog}. ChatStitch accurately performed perspective adjustment and localization of occluded objects, whereas VisProg consistently produced incorrect results. 
The reason for this discrepancy is that VisProg does not inherently support perspective, and is unable to obtain information about occluded objects.ct instructions.

\subsubsection{Stitching Results}
\label{sec:stitching-results}

We conducted extensive experiments on the UDIS-D\cite{UDIS} dataset and the MCOV-SLAM\cite{PanMiaoxin} dataset to demonstrate the performance of our proposed stitching framework, especially its ability to handle multi-image stitching tasks under the non-global-overlapping condition.

\textbf{Quantitative comparison.}
We first conduct multi-image stitching comparison with UDIS2\cite{UDIS2} on the two datasets, calculating PSNR and SSIM metrics to compare the image warping and alignment performance.
The results are shown in \cref{tab:comparison-multi-images}, where $I_{3\times3}$ represents the identity matrix, indicating no warping or alignment of the target image, serving as a reference.
The evaluation results are divided into three levels based on performance, as described in UDIS\cite{UDIS} and UDIS2\cite{UDIS2}.
Next, we conduct two-image stitching comparison with other classic methods\cite{APAP-TPAMI, SPW, LPC, UDIS} on these two datasets. The results are shown in \cref{tab:comparison-two-images}.

\begin{figure*}[ht]
  \centering
   \includegraphics[width=\linewidth]{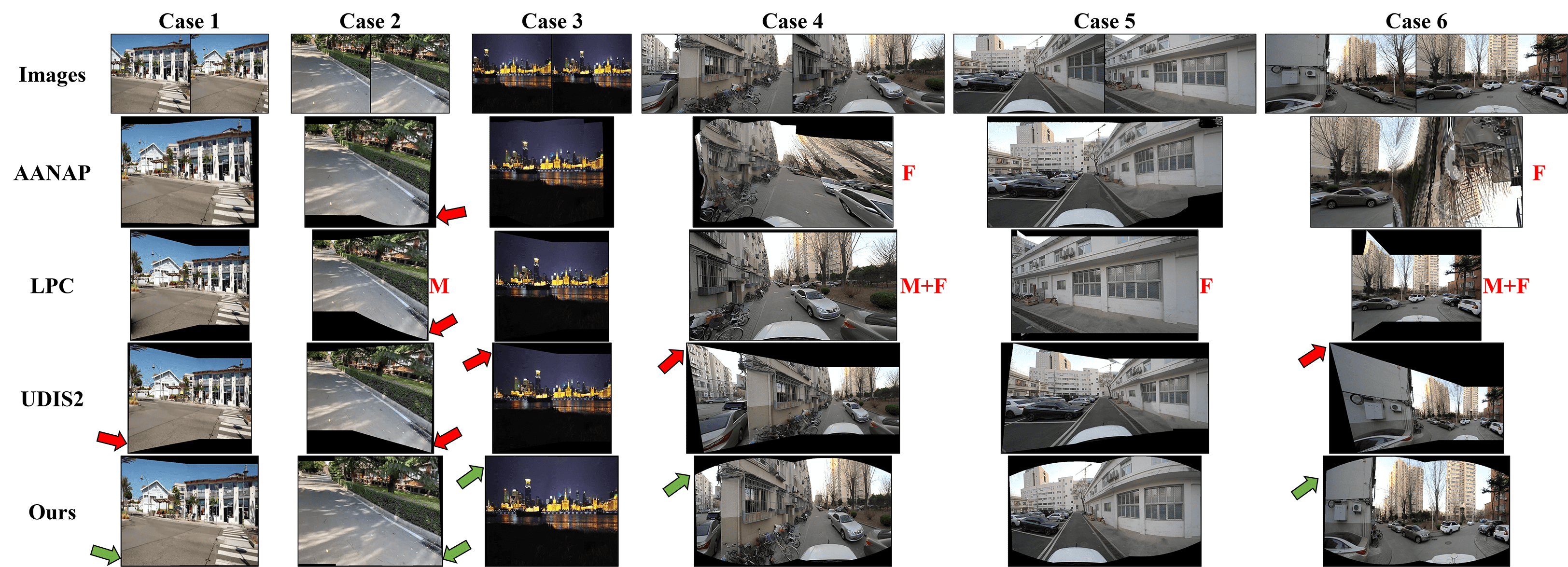}
   \caption{Qualitative comparison on the UDIS-D\cite{UDIS} and MCOV-SLAM\cite{PanMiaoxin} datasets with the AANAP\cite{AANAP}, LPC\cite{LPC}, and UDIS2\cite{UDIS2} methods. Case 1-2 are from UDIS-D, while Case 3-5 are from MCOV-SLAM. \textbf{\textcolor{red}{F}} indicates cases of stitching failure. \textbf{\textcolor{red}{M}} indicates that the LPC\cite{LPC} algorithm cannot extract effective line feature matches and requires manual annotation of line feature matches.}
   \label{fig:two-stitch}
   \vspace{-10pt}
\end{figure*}   

\begin{figure*}[ht]
  \centering
   \includegraphics[width=\linewidth]{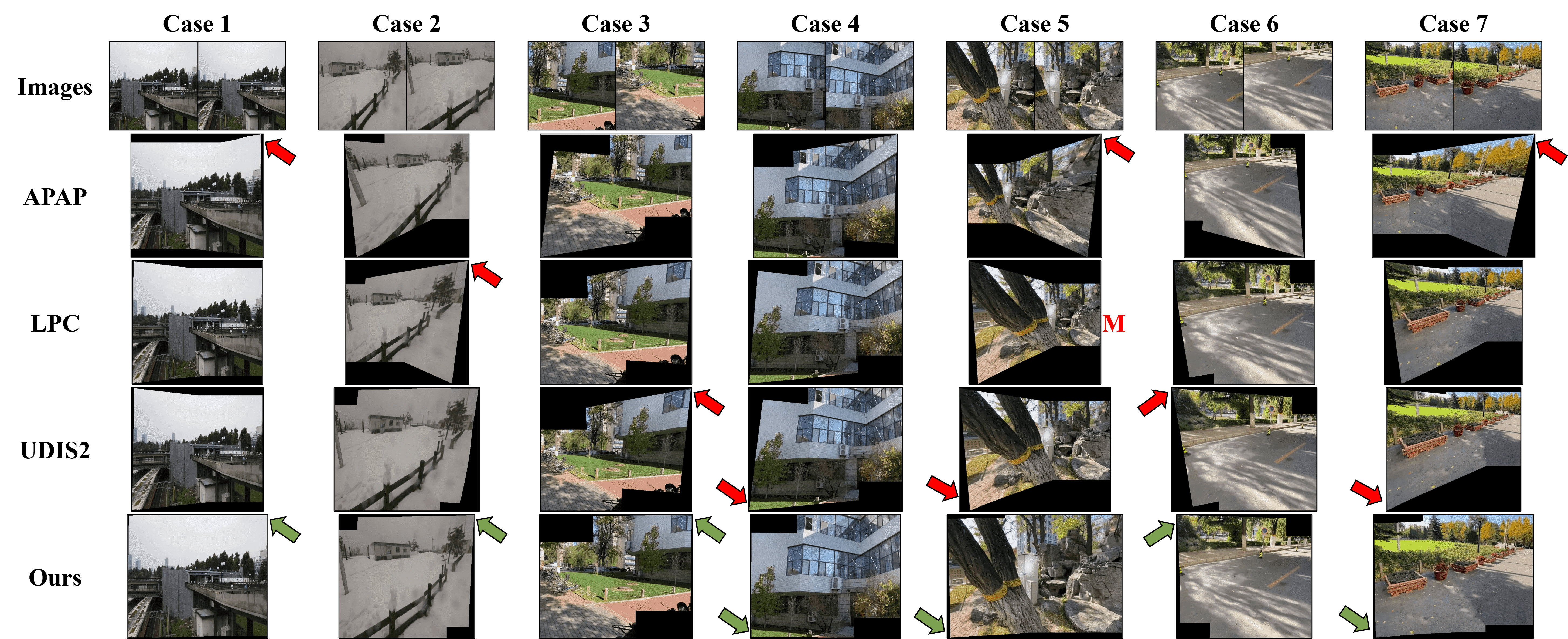}
   \caption{More qualitative comparison of two-image stitching on the UDIS-D\cite{UDIS} dataset with the APAP\cite{APAP-TPAMI}, LPC\cite{LPC}, and UDIS2\cite{UDIS2} methods. \textbf{\textcolor{red}{M}} indicates that the LPC\cite{LPC} algorithm cannot extract effective line feature matches and requires manual annotation of line feature matches.}
   \label{fig:suppl-two-stitch-udis}
   \vspace{-20pt}
\end{figure*}

\begin{figure*}[ht]
  \centering
   \includegraphics[width=\linewidth]{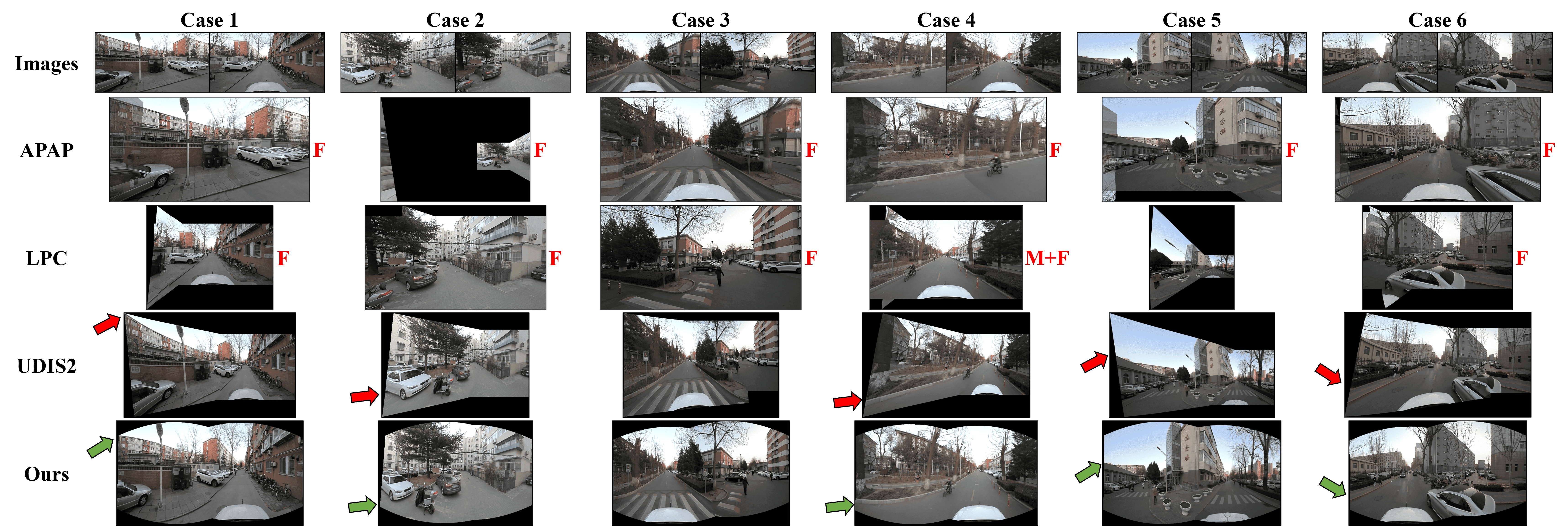}
   \caption{More qualitative comparison of two-image stitching on the MCOV-SLAM\cite{PanMiaoxin} dataset with the APAP\cite{APAP-TPAMI}, LPC\cite{LPC}, and UDIS2\cite{UDIS2} methods. \textbf{\textcolor{red}{F}} indicates cases of stitching failure. \textbf{\textcolor{red}{M}} indicates that the LPC\cite{LPC} algorithm cannot extract effective line feature matches and requires manual annotation of line feature matches.}
   \label{fig:suppl-two-stitch-mcov}
   \vspace{-20pt}
\end{figure*}

\begin{table}
\caption{Quantitative comparison of two-image stitching  on the UDIS-D\cite{UDIS} and MCOV-SLAM\cite{PanMiaoxin} datasets. E, M, H, and A denote Easy, Moderate, Hard, and Average, respectively. \textcolor{red}{Red} indicates the best results.}
\centering
\resizebox{\linewidth}{!}{
\begin{tabular}{cccccc|cccc}
\toprule
\multirow{2}*{Datasets} &  \multirow{2}*{Methods} &
    \multicolumn{4}{c}{PSNR$\uparrow$} & \multicolumn{4}{c}{SSIM$\uparrow$} \\
    \cmidrule(r){3-6} \cmidrule(r){7-10}
    & & E & M & H & A & E & M & H & A \\
\midrule
\multirow{6}*{\makecell[c]{UDIS\\-D\cite{UDIS}}} 
    & $I_{3\times3}$ & 15.87 & 12.76 & 10.68 & 12.86 & 0.530 & 0.286 & 0.146 & 0.303 \\
    & APAP\cite{APAP-TPAMI} & 27.96 & 24.39 & 20.21 & 23.79 & 0.901 & 0.837 & 0.682 & 0.794 \\
    & SPW\cite{SPW} & 26.98 & 22.67 & 16.77 & 21.60 & 0.880 & 0.758 & 0.490 & 0.687 \\
    & LPC\cite{LPC} & 26.94 & 22.63 & 19.31 & 22.59 & 0.878 & 0.764 & 0.610 & 0.736 \\
    & UDIS\cite{UDIS} & 25.16 & 20.96 & 18.36 & 21.17 & 0.834 & 0.669 & 0.495 & 0.648 \\
    & Ours & \textcolor{red}{28.79} & \textcolor{red}{24.57} & \textcolor{red}{20.71} & \textcolor{red}{24.28} & \textcolor{red}{0.914} & \textcolor{red}{0.841} & \textcolor{red}{0.692} & \textcolor{red}{0.803} \\
\midrule
\multirow{3}*{\makecell[c]{MCOV-\\SLAM\\\cite{PanMiaoxin}}} 
    & $I_{3\times3}$ & 12.59 & 11.33 & 10.09 & 11.21 & 0.356 & 0.264 & 0.175 & 0.256 \\
    & UDIS\cite{UDIS} & \textcolor{red}{18.03} & \textcolor{red}{16.17} & 13.37 & 15.60 & 0.674 & 0.561 & 0.320 & 0.498 \\
    & Ours & 17.32 & 15.80 & \textcolor{red}{14.35} & \textcolor{red}{15.67} & \textcolor{red}{0.751} & \textcolor{red}{0.694} & \textcolor{red}{0.555} & \textcolor{red}{0.655} \\
\bottomrule
\end{tabular}
}
\label{tab:comparison-two-images}
\end{table}

\textbf{Qualitative comparison.}
\cref{fig:two-stitch} shows the qualitative results of two-image stitching on the two datasets. It can be seen that even at extremely low overlap rates, our method can still smoothly warp the target image.
\cref{fig:suppl-two-stitch-udis} and \cref{fig:suppl-two-stitch-mcov} respectively show the comparison of two-image stitching results on the UDIS-D\cite{UDIS} and MCOV-SLAM\cite{PanMiaoxin} datasets.
Traditional methods APAP\cite{APAP-TPAMI} and LPC\cite{LPC} experienced a large number of stitching failures on the MCOV-SLAM dataset.
Although UDIS2\cite{UDIS2} can achieve good alignment results on both datasets, there is serious projective distortion, leading to severe stretching in the warped images.
In contrast, our method can achieve precise alignment while reducing the projective distortion of the warped image, providing a solid foundation for multi-image stitching.

\begin{figure}[t]
  \centering
   \includegraphics[width=\linewidth]{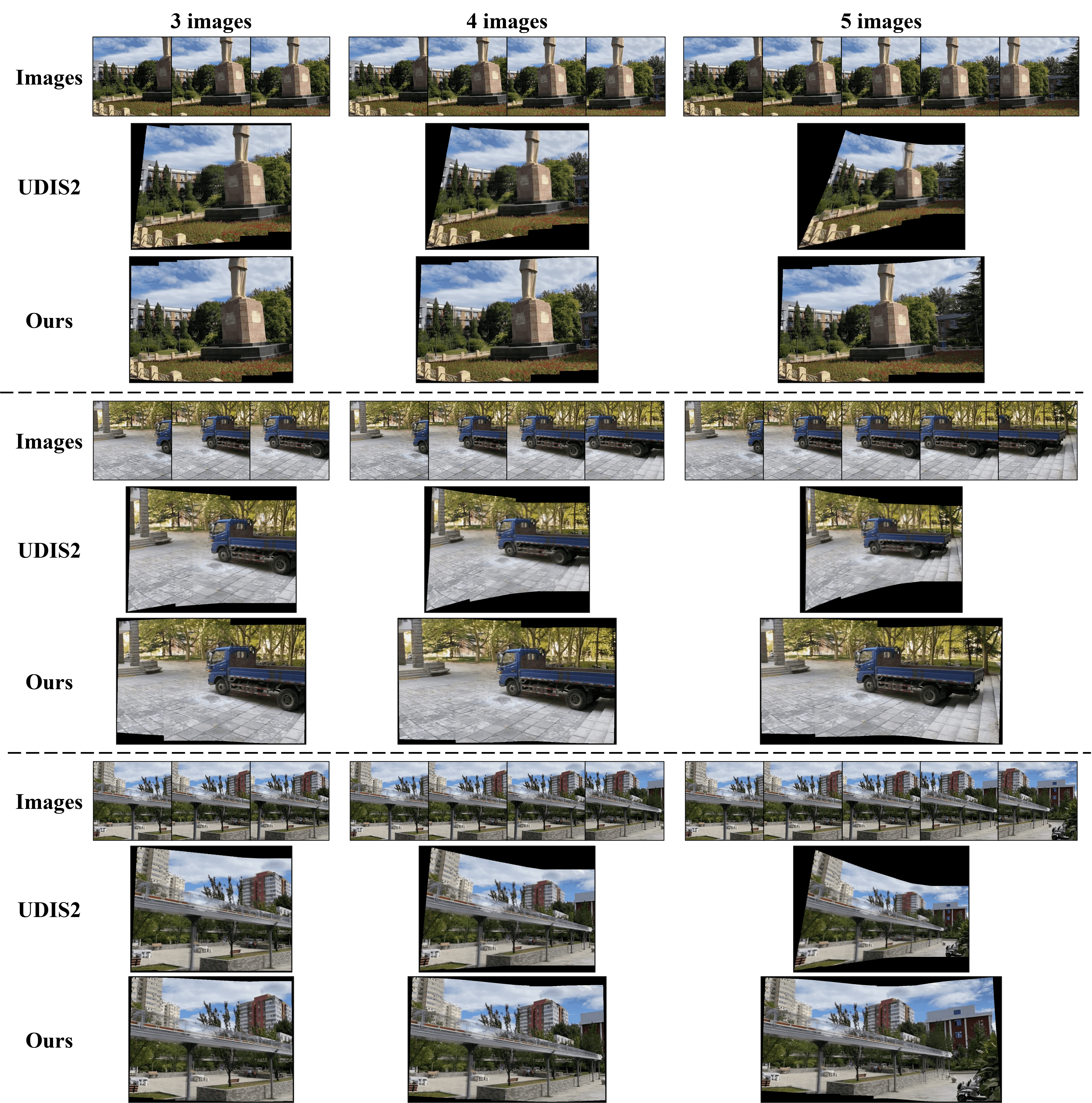}
   \caption{More qualitative comparison of multi-image stitching on the UDIS-D\cite{UDIS} dataset with the UDIS2\cite{UDIS2} methods.}
   \label{fig:suppl-multi-stitch-udis}
\end{figure}

\begin{figure}[ht]
  \centering
   \includegraphics[width=\linewidth]{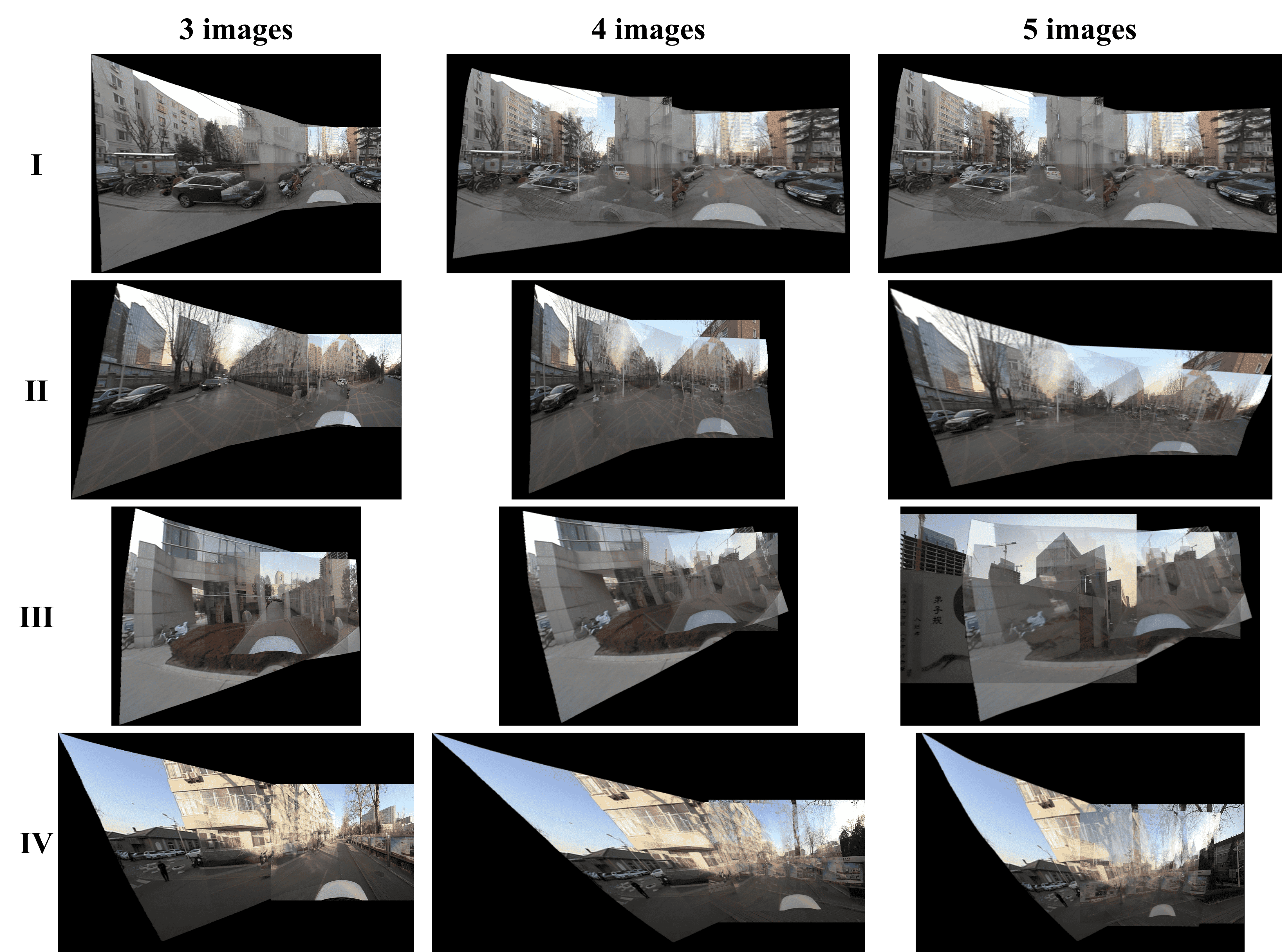}
   \caption{The results of multi-image stitching by UDIS2\cite{UDIS2} on the MCOV-SLAM\cite{PanMiaoxin} dataset.}
   \label{fig:suppl-multi-stitch-fail}
\end{figure}

\begin{figure*}[ht]
  \centering
   \includegraphics[width=\linewidth]{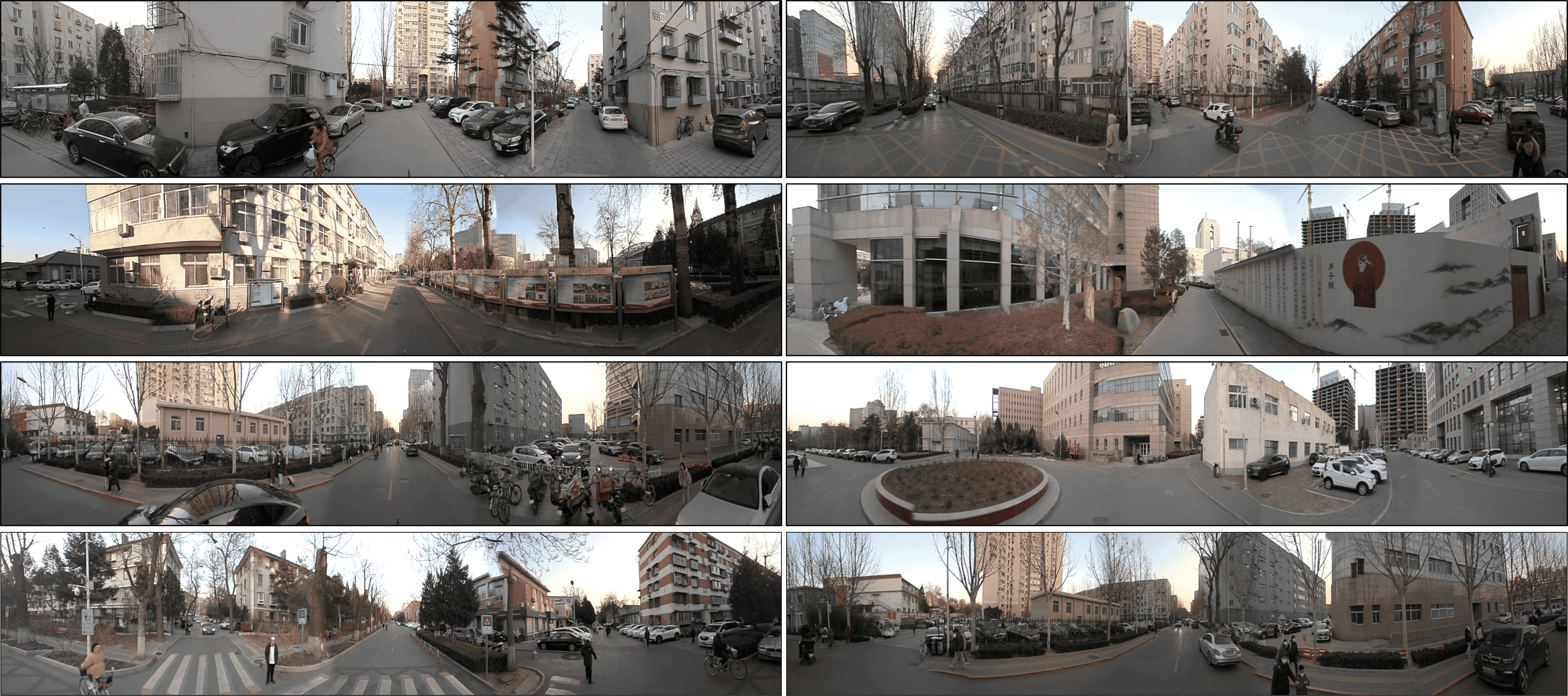}
   \caption{The results of multi-image stitching by our SV-UDIS on the MCOV-SLAM\cite{PanMiaoxin} dataset.}
   \label{fig:suppl-multi-stitch-results}
\end{figure*}

\cref{fig:suppl-multi-stitch-udis} shows a comparison of multi-image stitching results on the UDIS-D\cite{UDIS} dataset using our method and UDIS2\cite{UDIS2}.
As can be seen, our method, through the motion propagation and rectangular warping constraints, can effectively extend two-image stitching to multi-image stitching, ensuring alignment accuracy while reducing projective distortion.

\cref{fig:suppl-multi-stitch-fail} shows the multi-image stitching results of UDIS2\cite{UDIS2} on the MOCV-SLAM\cite{PanMiaoxin} dataset. When stitching three images, UDIS2 already shows signs of stitching failure. When stitching 4 or 5 images, UDIS2 almost completely fails to stitch.
In contrast, \cref{fig:suppl-multi-stitch-results} shows the results of stitching five images on the MCOV-SLAM dataset using our proposed SV-UDIS.

\textbf{Ablation studies.}
We investigated the impact of cylindrical projection and rectangular warping constraints on the effects of two-image and multi-image stitching. 

\begin{figure}[ht]
  \centering
   \includegraphics[width=\linewidth]{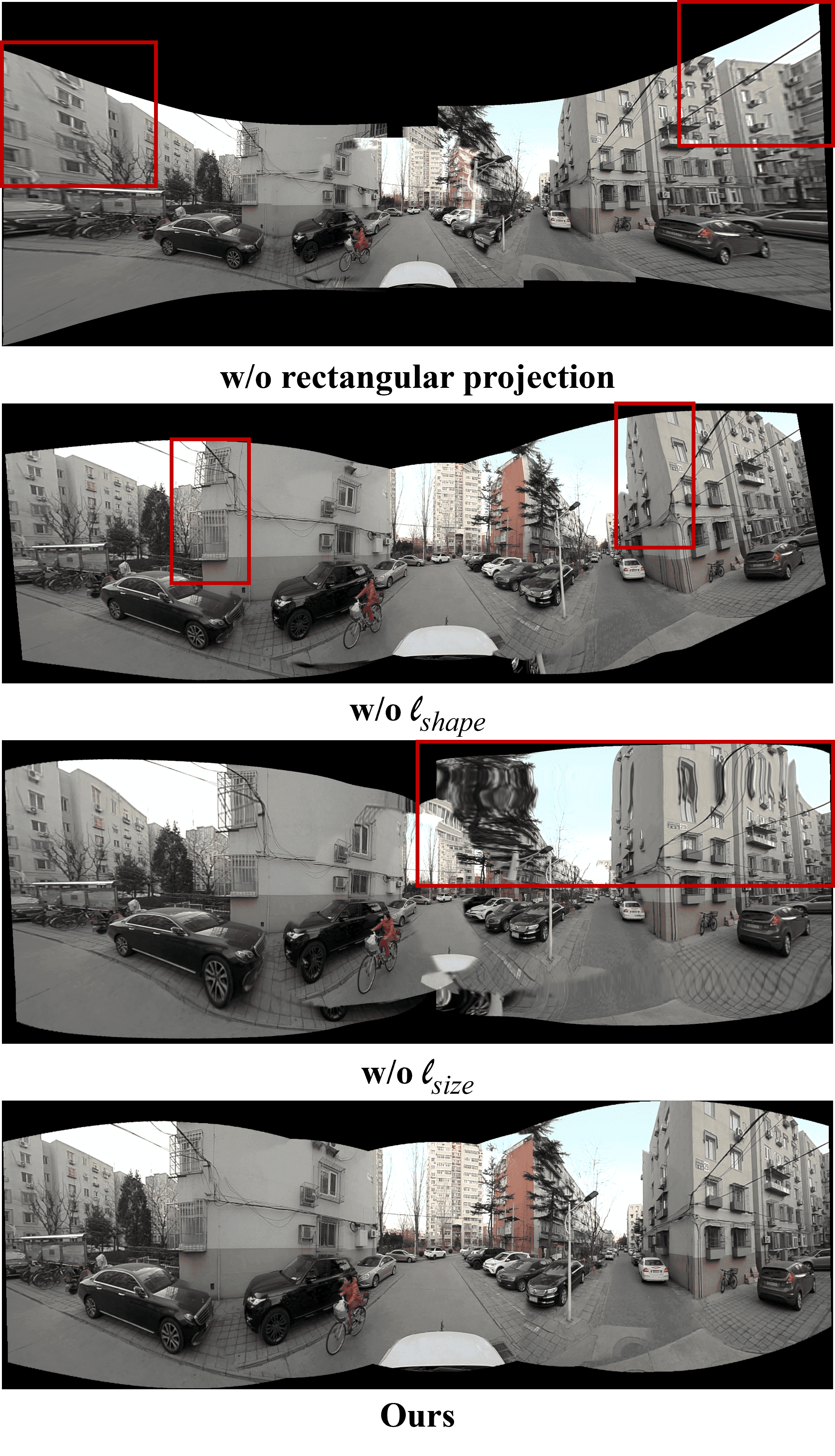}
   \caption{Ablation studies on cylindrical projection and our rectangular warping constraints.}
   \label{fig:suppl-ablation-stitch}
\end{figure}

We conduct ablation studies on cylindrical projection and different rectangular warping constraints.
The quantitative results and qualitative results are shown in \cref{tab:suppl-ablation-stitch} and \cref{fig:suppl-ablation-stitch}, respectively.
From these, it can be seen that:
\begin{itemize}
    \item Cylindrical projection can effectively improve the alignment accuracy on the MCOV-SLAM\cite{PanMiaoxin} dataset, and can significantly alleviate the projective distortion when stitching multiple images.
    \item The rectangular shape constraint $\ell_{shape}$ may reduce the alignment accuracy, but it can effectively alleviate the geometric structure distortion and skewness when stitching multiple images.
    \item The rectangular size constraint $\ell_{size}$ may slightly reduce the alignment accuracy, but it can effectively alleviate the blurring of the image caused by drastic changes in grid size.
    \item The rectangular fold constraint $\ell_{fold}$ can slightly improve the alignment accuracy, while avoiding the grid folding phenomenon.
\end{itemize}

\begin{table}
\centering
\caption{Ablation studies on the UDIS-D\cite{UDIS} and MCOV-SLAM\cite{PanMiaoxin} datasets. C.p. denotes cylindrical projection.}
\resizebox{\linewidth}{!}{
\begin{tabular}{cccccc|cccc}
\toprule
\multirow{2}*{Datasets} &  \multirow{2}*{Methods} &
    \multicolumn{2}{c}{2 images} & \multicolumn{2}{c}{3 images} & \multicolumn{2}{c}{4 images} & \multicolumn{2}{c}{5 images} \\
    \cmidrule(r){3-4} \cmidrule(r){5-6} \cmidrule(r){7-8} \cmidrule(r){9-10}
    & & PSNR & SSIM & PSNR & SSIM & PSNR & SSIM & PSNR & SSIM \\
\midrule
\multirow{4}*{\makecell[c]{UDIS\\-D$^\dag$\cite{UDIS}}} 
    & w/o $\ell_{shape}$         & 24.60 & 0.820 & 23.95 & 0.794 & 24.17 & 0.803 & 24.09 & 0.804 \\
    & w/o $\ell_{size}$          & 24.41 & 0.807 & 23.74 & 0.786 & 24.03 & 0.796 & 23.94 & 0.797 \\
    & w/o $\ell_{fold}$          & 24.10 & 0.796 & 23.44 & 0.777 & 23.62 & 0.783 & 23.52 & 0.784 \\
    & Ours                       & 24.28 & 0.803 & 23.58 & 0.784 & 23.77 & 0.790 & 23.62 & 0.789 \\
\midrule
\multirow{5}*{\makecell[c]{MCOV-\\SLAM\\\cite{PanMiaoxin}}} 
    & w/o c.p. & 15.11 & 0.592 & 15.39 & 0.572 & 15.28 & 0.595 & 15.10 & 0.572 \\
    & w/o $\ell_{shape}$         & 15.81 & 0.662 & 15.87 & 0.649 & 15.77 & 0.665 & 15.59 & 0.638 \\
    & w/o $\ell_{size}$          & 13.92 & 0.549 & 14.21 & 0.540 & 13.97 & 0.561 & 13.83 & 0.528 \\
    & w/o $\ell_{fold}$          & 15.66 & 0.654 & 15.24 & 0.632 & 15.72 & 0.653 & 15.35 & 0.619 \\
    & Ours                       & 15.67 & 0.655 & 15.31 & 0.634 & 15.73 & 0.655 & 15.36 & 0.619 \\
\bottomrule
\end{tabular}
}
\label{tab:suppl-ablation-stitch}
\vspace{-15pt}
\end{table}

\section{Conclusion}
\label{sec:conclusion}
This paper introduces ChatStitch, a surround-view human-machine co-perception system capable of unveiling obscured blind spot information through natural language commands integrated with external digital assets. To dismantle the unidirectional interaction bottleneck, ChatStitch implements a cognitively grounded closed-loop interaction multi-agent framework based on LLMs. To suppress distortion propagation across overlapping boundaries, We propose SV-UDIS within ChatStitch, a technical module corresponding to an agent in the ChatStitch framework, designed to stitch multiple surround-view images captured in driving scenarios into a seamless panoramic image. Experiments demonstrate that ChatStitch can generate photorealistic surround-view perception outcomes that effectively respond to a variety of human language commands, enabling a perspective view of occluded objects. In the future, we will focus on integrating more perception functionalities such as object detection and depth estimation.

\bibliographystyle{IEEEtran}
\bibliography{main}

\begin{IEEEbiography}[{\includegraphics[width=1in,height=1.25in,clip,keepaspectratio]{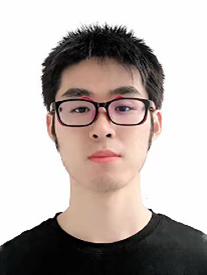}}]{Hao Liang}
received the B.S. degree in Beijing Institute of Technology, Beijing, China, in 2019. He is currently pursuing the Ph.D. degree with Integrated Navigation and Intelligent Navigation Laboratory, Beijing Institute of Technology. His research interests include computer vision and computer graphics. 
\end{IEEEbiography}
\begin{IEEEbiography}[{\includegraphics[width=1in,height=1.25in,clip,keepaspectratio]{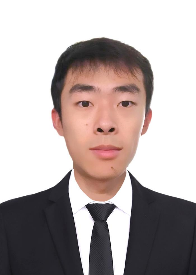}}]{Zhipeng Dong} received the B.S. degree in Beijing Institute of Technology, Beijing, China, in 2022. He is currently pursuing the Ph.D. degree with Integrated Navigation and Intelligent Navigation Laboratory, Beijing Institute of Technology. His research interests include computer vision and computer graphics. 
\end{IEEEbiography}
\begin{IEEEbiography}[{\includegraphics[width=1in,height=1.25in,clip,keepaspectratio]{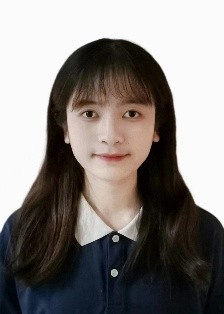}}]{Kaixin Chen} received the B.S. degree from  Beijing Institute of Technology, Beijing, China, in 2023. She is currently pursuing the M.S. degree with Integrated Navigation and Intelligent Navigation Laboratory, Beijing Institute of Technology. Her research interests include multi-view collaborative perception and image processing. 
\end{IEEEbiography} 
\begin{IEEEbiography}[{\includegraphics[width=1in,height=1.25in,clip,keepaspectratio]{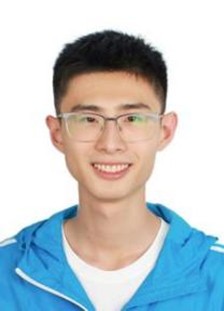}}]{Jiyuan Guo}  received his bachelor's and master's degrees from China University of Geosciences (Beijing) in 2021 and 2024, and is currently a Ph.D. student at Beijing Institute of Technology. His research focuses on image stitching and fusion. 
\end{IEEEbiography}
\begin{IEEEbiography}[{\includegraphics[width=1in,height=1.25in,clip,keepaspectratio]{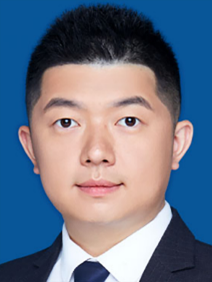}}]{Yufeng Yue} received the B.Eng. degree in automation from the Beijing Institute of Technology, Beijing, China, in 2014, the Ph.D. degree from Nanyang Technological University, Singapore, in 2019. He is currently a Professor with the School of Automation, Beijing Institute of Technology. He was a Visiting Scholar with the University of California, Los Angeles, CA, USA, in 2019. His research interests include perception, navigation, and coordination for multi-robot system in complex environments. He is a member of the IEEE.. 
\end{IEEEbiography}
\begin{IEEEbiography}[{\includegraphics[width=1in,height=1.25in,clip,keepaspectratio]{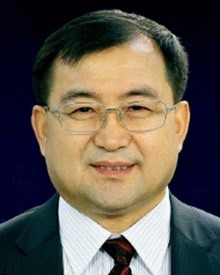}}]{Mengyin Fu} is currently a Part-Time Doctoral supervisor with the School of Automation, Beijing Institute of Technology, and a Professor with the School of Automation, Nanjing University of Science and Technology. His research interests include integrated navigation, intelligent navigation, image processing, learning, and recognition, as well as their applications. Dr. Fu was elected as the Academician of the Chinese Academy of Engineering in 2021. In recent years, he received the National Science and Technology Progress Award several times.  
\end{IEEEbiography}
\begin{IEEEbiography}[{\includegraphics[width=1in,height=1.25in,clip,keepaspectratio]{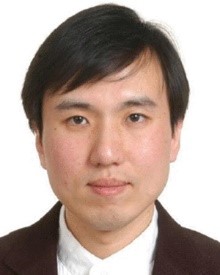}}]{Yi Yang} received the B.Eng. and M.Eng. degrees from the Hebei University of Technology, Tianjin, China, in 2001 and 2004, respectively, and the Ph.D. degree from the Beijing Institute of Technology, Beijing, China, in 2010. He is currently a Professor with the School of Automation, Beijing Institute of Technology. His research interests include robotics, autonomous systems, intelligent navigation, cross-domain collaborative perception, and motion planning and control. Dr. Yang received the National Science and Technology Progress Award twice. 
\end{IEEEbiography}
\end{document}